\documentclass[letterpaper,journal]{IEEEtran}

\usepackage{amsmath,amsfonts,amssymb}
\usepackage{array}
\usepackage{graphicx}
\usepackage{xspace}
\usepackage{booktabs}
\usepackage{multirow}
\usepackage{makecell}
\usepackage{adjustbox}
\usepackage{pifont}
\usepackage[table]{xcolor}
\usepackage{placeins}
\usepackage[numbers,sort&compress]{natbib}
\usepackage[hidelinks]{hyperref}
\usepackage[capitalize]{cleveref}
\usepackage{orcidlink}

\newcommand{\xmark}{\ding{55}}
\definecolor{highlightgray}{gray}{0.92}

\makeatletter
\DeclareRobustCommand\onedot{\futurelet\@let@token\@onedot}
\def\@onedot{\ifx\@let@token.\else.\null\fi\xspace}
\makeatother

\def\eg{\emph{e.g}\onedot}
\def\cf{\emph{cf}\onedot}

\newcommand{\model}{\emph{{STAMP}}\xspace}
\newcommand{\modelP}{\emph{{STAMPlus}}\xspace}

\begin{document}

\title{Better, Stronger, Faster, and Broader: Structured All-Mask Prediction for MLLM-Based Segmentation}

\author{Jiazhen Liu\,\orcidlink{0000-0003-0584-4571},
Mingkuan Feng\,\orcidlink{0009-0000-9282-0695}, and
Long Chen\,\orcidlink{0000-0001-6148-9709}\,\href{mailto:longchen@ust.hk}{\textsuperscript{\ding{41}}}
\thanks{The authors are with The Hong Kong University of Science and Technology (HKUST), Hong Kong.}
\thanks{A preliminary version of this work was presented at CVPR 2026~\cite{liu2026STAMP}.}
\thanks{Long Chen is the corresponding author (e-mail: \href{mailto:longchen@ust.hk}{longchen@ust.hk}).}}

\markboth{Submission}
{Liu \MakeLowercase{\textit{et al.}}: Better, Stronger, Faster, and Broader}

\maketitle

\begin{abstract}

MLLM-based segmentation faces a core \textbf{segmentation trilemma}: high segmentation performance, preserved dialogue ability, and fast inference. Embedding-prediction methods may disrupt language modeling through pixel-level objectives, whereas next-token generation is inefficient for dense masks. We propose \textbf{All-Mask Prediction}, decoupling autoregressive dialogue from non-autoregressive mask prediction. Its binary instantiation, \model{} (\textbf{S}imultaneous \textbf{T}extual \textbf{A}ll-\textbf{M}ask \textbf{P}rediction), emits an in-vocabulary \texttt{<SEG>} trigger, fuses image-aligned mask tokens with corresponding patch features, and uses hybrid attention to classify all tokens as foreground or background in one pass. It thereby combines strong referring and reasoning segmentation with preserved multimodal ability and efficient inference. However, binary masks cannot retain multiple semantic or instance identities without repeated target-specific predictions.
We therefore propose \textbf{Structured All-Mask Prediction} and develop \modelP{}. It generates a target list with explicit IDs and optional boxes, binds these IDs to a shared multi-class mask space, and jointly predicts all targets in one non-autoregressive pass. A single unified checkpoint retains \model{}'s referring and reasoning capabilities while extending to open-vocabulary semantic, instance-aware, and remote-sensing small-target segmentation, where high-resolution mask-token scaling preserves finer spatial evidence.
Across these settings, \modelP{} achieves state-of-the-art segmentation performance, preserves general multimodal instruction following, and reduces 12-category latency from 13.50s for repeated \model{} inference to 5.16s. Further analyses show that accurate target cues improve segmentation and learned spatial grounding benefits look-twice reasoning. Overall, \modelP{} resolves the trilemma beyond single-target prediction. The complete codebase is included in the supplementary material.

\end{abstract}

\section{Introduction}
\label{sec:intro}

The success of Multimodal Large Language Models (MLLMs) has spurred a trend to unify diverse vision tasks within a single instruction-driven framework~\cite{yang2025qwen3, chen2024lion, zhang2025mllms, jiang2025detect, zhu2025internvl3, li2024llava}. 
For such models to be practical, they should simultaneously (i) preserve dialogue ability, (ii) achieve high task performance, and (iii) maintain fast inference. 
While this goal has been increasingly realized for recognition~\cite{zhu2025internvl3, li2024llava, liu2025phd} and detection~\cite{bai2025qwen2, yang2025qwen3}, it remains challenging for segmentation, where models must produce dense, pixel-level masks. 
This difficulty stems from a fundamental mismatch: the sequential text-generative nature of MLLMs is ill-suited for dense pixel generation~\cite{liu2025segmentation, lan2024text4seg}. 
Consequently, even when segmenting a single language-specified target, current MLLM-based segmentation methods are forced into a core trilemma, where they must compromise on one or more of these fronts.

\begin{figure}[t]
    \centering
    \includegraphics[width=0.75\linewidth]{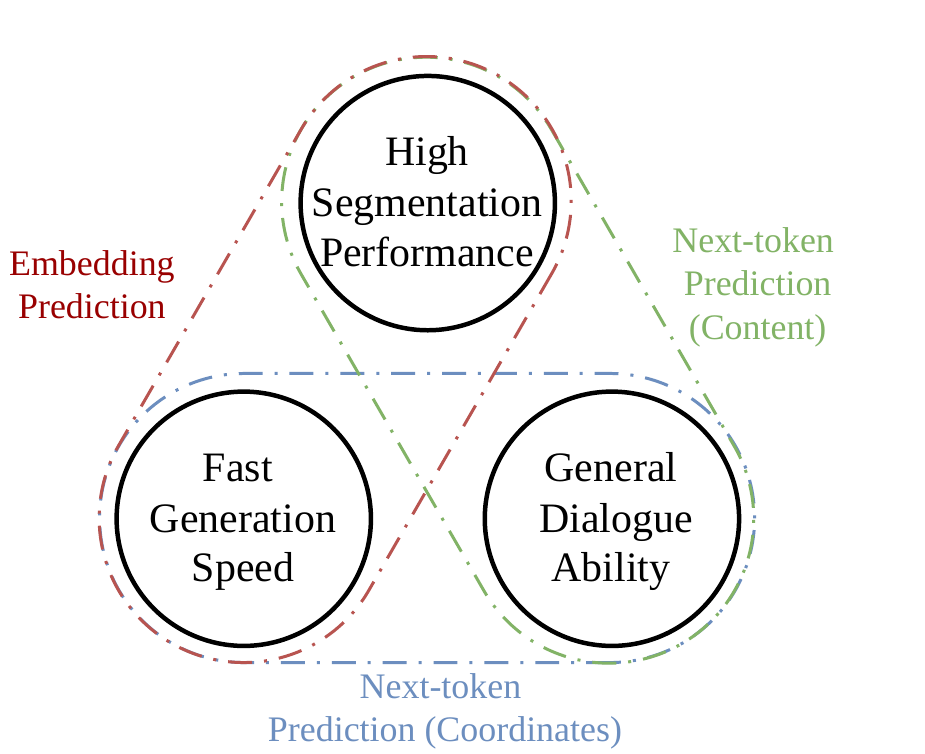}
    \caption{\textbf{The trilemma of segmentation in MLLMs.} Embedding prediction may harm dialogue abilities. Next-token prediction methods are either fast with poor segmentation performance or achieve superior performance at the cost of inference speed, particularly when generating rich content (\eg, chain-of-thought or patch-wise classification).}
    \label{fig:tri_problem}
\end{figure}

\begin{table}[t]
    \centering
    \scriptsize
    \renewcommand{\arraystretch}{1.02}
    \caption{\textbf{Comparison of MLLM segmentation paradigms.} ``Mask Steps'' counts the mask-generation steps for $N$ targets, excluding textual prefixes. ``Token-only Supv.'' indicates supervision within the token interface, and ``Decoder-free'' indicates that no external mask decoder is required.}
    
    \setlength{\tabcolsep}{1.5pt}
    \begin{tabular}{@{}
        >{\raggedright\arraybackslash}p{0.27\columnwidth}
        >{\raggedright\arraybackslash}p{0.19\columnwidth}
        >{\centering\arraybackslash}p{0.25\columnwidth}
        cc@{}}
        \toprule
        \textbf{Model} 
        & \textbf{\makecell[l]{Output}} 
        & \textbf{\makecell[r]{Mask Steps\\for $N$ Targets}} 
        & \textbf{\makecell[r]{Token-only\\Supv.}} 
        & \textbf{\makecell[r]{Decoder-\\free}} \\
        \midrule
        
        \multicolumn{5}{l}{\textit{\textbf{Paradigm 1: Embedding Prediction}}} \\
        LISA {\tiny (CVPR'24)}~\cite{lai2024lisa} 
        & Embeddings 
        & $\mathcal{O}(N)$ 
        & \xmark 
        & \xmark \\
        
        GSVA {\tiny (CVPR'24)}~\cite{xia2024gsva} 
        & Embeddings 
        & $\mathcal{O}(N)$ 
        & \xmark 
        & \xmark \\
        
        PixelLM {\tiny (CVPR'24)}~\cite{ren2024pixellm} 
        & Embeddings 
        & $\mathcal{O}(N)$ 
        & \xmark 
        & \xmark \\
        
        M$^2$SA {\tiny (ICLR'25)}~\cite{jang2025mmr} 
        & Embeddings 
        & $\mathcal{O}(N)$ 
        & \xmark 
        & \xmark \\
        
        READ {\tiny (CVPR'25)}~\cite{read} 
        & Embeddings 
        & $\mathcal{O}(N)$ 
        & \xmark 
        & \xmark \\
        \midrule

        \multicolumn{5}{l}{\textit{\textbf{Paradigm 2: Next-token Prediction}}} \\
        VisionLLM {\tiny (NIPS'24)}~\cite{wu2024visionllm} 
        & Coordinates 
        & $\mathcal{O}(N \times N_{\text{points}})$ 
        & \checkmark 
        & \checkmark \\
        
        Seg-Zero {\tiny (arXiv'25)}~\cite{liu2025seg} 
        & CoT + Coords. 
        & $\mathcal{O}(N \times N_{\text{CoT}})$ 
        & \xmark 
        & \xmark \\
        
        SegAgent {\tiny (CVPR'25)}~\cite{zhu2025segagent} 
        & CoT + Coords. 
        & $\mathcal{O}(N \times N_{\text{CoT}})$ 
        & \checkmark 
        & \xmark \\
        
        Text4Seg {\tiny (ICLR'25)}~\cite{lan2024text4seg} 
        & Patch Class. 
        & $\mathcal{O}(N \times N_{\text{patches}})$ 
        & \checkmark 
        & \checkmark \\

        Text4Seg++ {\tiny (TPAMI'26)}~\cite{text4segpp} 
        & Patch Class. 
        & $\mathcal{O}(N \times N_{\text{bbox}})$ 
        & \checkmark 
        & \checkmark \\
        
        \midrule

        \multicolumn{5}{l}{\textit{\textbf{Paradigm 3: All-Mask Prediction (Ours)}}} \\
        \rowcolor{highlightgray}
        \model (Ours)~\cite{liu2026STAMP}
        & Patch Class. 
        & $\mathcal{O}(N)$ 
        & \checkmark 
        & \checkmark \\
        
        \rowcolor{highlightgray}
        \modelP (Ours)
        & Patch Class. 
        & $\mathcal{O}(1)$ 
        & \checkmark 
        & \checkmark \\

        \bottomrule
    \end{tabular}
    \label{tab:comp-paradigm}
\end{table}

\begin{figure*}[t]
    \centering
    \includegraphics[width=\linewidth]{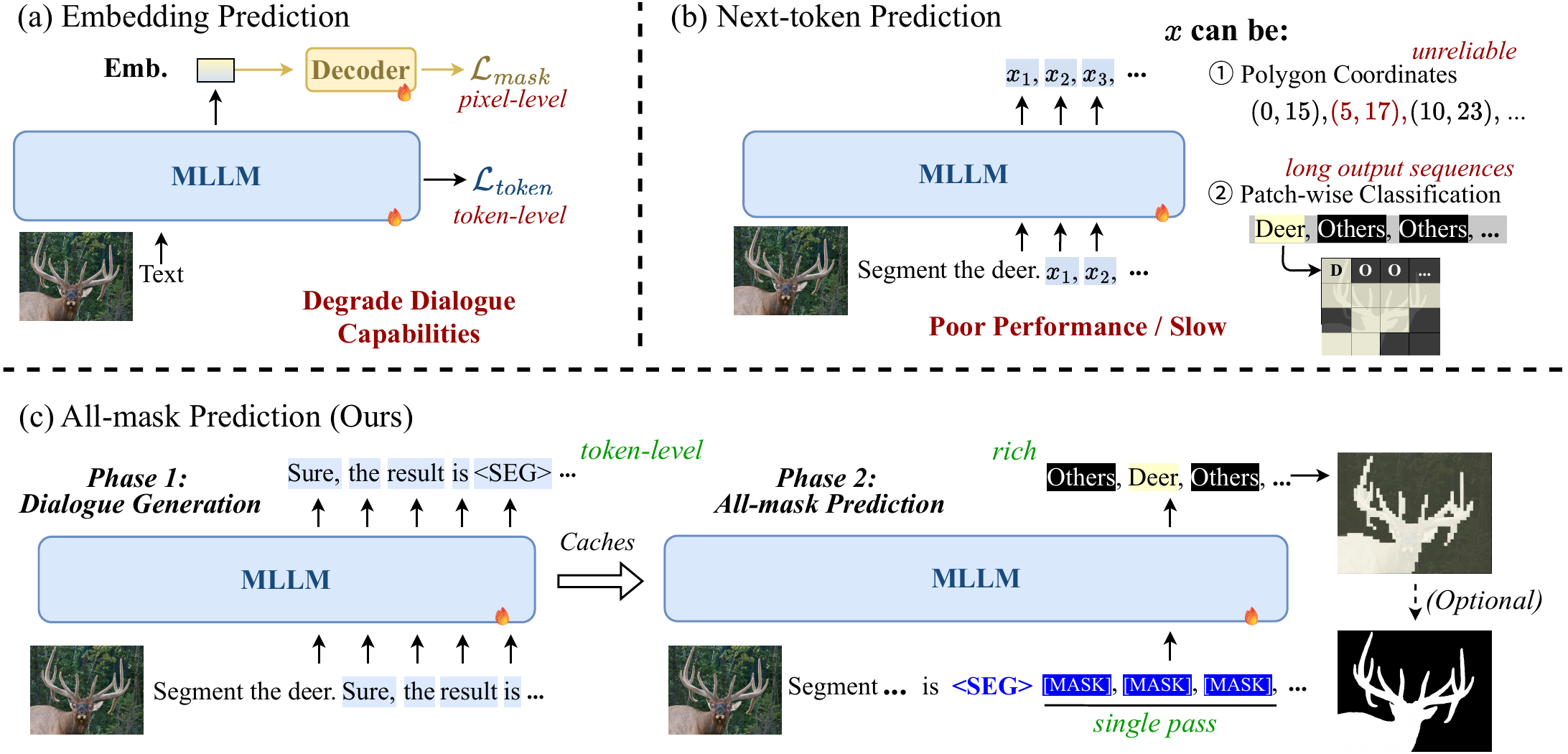}
\caption{\textbf{Comparison of MLLM-based segmentation paradigms.} 
\textbf{(a) Embedding Prediction}: Pixel-level supervision for external mask decoding~\cite{lai2024lisa, ren2024pixellm} may degrade the MLLM's general dialogue capabilities.
\textbf{(b) Next-Token Prediction}: Autoregressively generates textual mask representations~\cite{wang2023visionllm, liu2025seg, lan2024text4seg}, forcing a trade-off between poor segmentation performance (for sparse outputs) and slow inference (for rich outputs).
\textbf{(c) Our All-Mask Prediction}: We decouple dialogue generation (autoregressive) from mask generation (non-autoregressive). By simultaneously predicting all mask tokens as patch-wise classifications in a single pass, our paradigm resolves the segmentation trilemma, uniting preserved dialogue abilities, high segmentation performance and fast inference speed.
}
    \label{fig:intro}
\end{figure*}

\cref{fig:tri_problem} illustrates how mainstream MLLM segmentation paradigms navigate this trilemma, with their primary differences rooted in the MLLM's role and output format (\cref{tab:comp-paradigm}). The first paradigm, \emph{embedding prediction} (\cf \cref{fig:intro}a), achieves good segmentation performance by fine-tuning the MLLM with a pixel-level mask loss to produce target-specific embeddings, which drive an external binary-mask decoder to generate a mask for each referred target~\citep{lai2024lisa, ren2024pixellm, read, xia2024gsva, rasheed2024glamm}. However, this auxiliary pixel-level objective departs from the native token-generation interface and may degrade general dialogue abilities~\citep{lan2024text4seg, liu2025segmentation, wu2024see}. For example, LISA~\citep{lai2024lisa} may fail to follow a simple instruction like ``How many objects are there?'' and instead output a segmentation result~\cite{liu2025segmentation}.

The second paradigm, \emph{next-token prediction} (\cf \cref{fig:intro}b), avoids this objective conflict by reframing segmentation as a pure language modeling task, where the MLLM autoregressively generates a textual representation of the mask, such as coordinates~\citep{wang2023visionllm}, CoT-guided coordinates~\citep{liu2025seg, zhu2025segagent}, or patch-wise foreground/background labels~\cite{lan2024text4seg, text4segpp}. 
While richer textual representations improve segmentation quality, they require the model to autoregressively generate lengthy token sequences for each referred target, making dense mask prediction prohibitively slow.

Thus, existing paradigms remain constrained by the segmentation trilemma: they either compromise dialogue generality or preserve the language interface at the cost of inefficient autoregressive mask generation. To break this trilemma, we introduce \emph{All-Mask Prediction}, which decouples autoregressive dialogue generation from non-autoregressive mask generation (\cf \cref{fig:intro}c). The MLLM first follows standard autoregressive dialogue generation and then predicts all image-aligned mask placeholders simultaneously as patch-wise classifications in a single forward pass. This preserves the native token interface, provides compatible token-level supervision through patch-aligned mask tokens, and enables efficient non-autoregressive mask prediction.

In our initial work~\cite{liu2026STAMP}, we instantiated this paradigm as \emph{Binary All-Mask Prediction} and resolved the segmentation trilemma for single-target referring and reasoning segmentation~\cite{lai2024lisa, liu2025seg, text4segpp, zhu2025segagent, read, ren2024pixellm}.
Here, a target may be an individual object or the merged union of several referred regions; because these tasks do not require their internal semantic or instance identities to remain distinguishable, the output can be represented by a single binary foreground/background mask.

Binary All-Mask Prediction follows two phases.
In Phase~1, the MLLM autoregressively generates a free-form descriptive response ending with \texttt{<SEG>}.
This trigger starts Phase~2, where image-aligned \texttt{[MASK]} tokens are directly prefilled into the MLLM and processed in a single forward pass; a single linear binary-classification head maps their final hidden states to foreground/background labels.
Based on this paradigm, we developed \model{}.
\model{} achieves state-of-the-art results on the RefCOCO family~\cite{kazemzadeh2014referitgame,mao2016generation} and ReasonSeg~\cite{lai2024lisa}, while matching embedding-prediction methods in mask-generation speed and retaining the general multimodal performance of its Qwen2-VL backbone~\cite{wang2024qwen2}.
These results verify that \model{} resolves the accuracy--dialogue-compatibility--efficiency trilemma in this setting.

\begin{figure*}[t]
    \centering
    \includegraphics[width=0.8\textwidth]{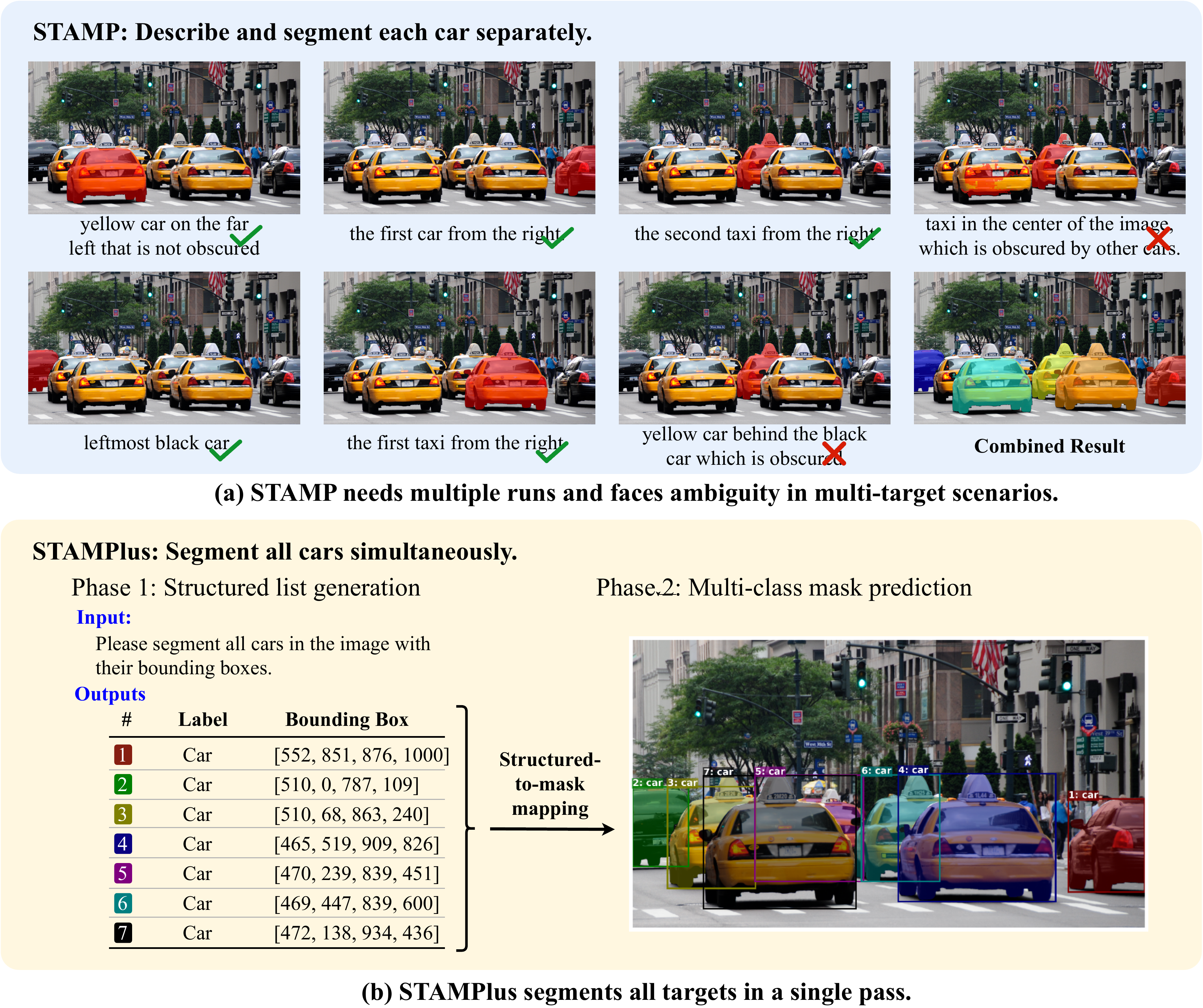}
    \caption{\textbf{\model{} versus \modelP{} in multi-target segmentation.}
    \textbf{(a)} \model{} must repeat its single-target prediction for each object, and ambiguous referring expressions may yield incorrect masks.
    \textbf{(b)} \modelP{} first generates a structured target list and then predicts all target IDs jointly in one shared structured mask map.}
    \label{fig:comp_with_pp}
\end{figure*}

Beyond the single-target setting, however, the task encompasses substantially more diverse scenarios, including open-vocabulary semantic, instance, and panoptic segmentation.
These tasks require not only delineating regions but also distinguishing their semantic categories or instance identities within a shared output.
\model{} and related binary-mask methods~\cite{lai2024lisa,ren2024pixellm,xia2024gsva,text4segpp,read} are therefore ill-suited to multi-target settings.
As shown in \cref{fig:comp_with_pp}a, they can produce multiple masks only by repeating the single-target process and merging its outputs, which sacrifices efficiency; moreover, the referring expressions needed to distinguish similar targets can be ambiguous.

Thus, the trilemma remains largely unaddressed in broader segmentation settings.
We therefore present an extension of our initial work to this broader regime.
We address this gap by generalizing All-Mask Prediction beyond binary target masks while retaining its original goals of accuracy, dialogue compatibility, and efficiency.
Such an extension naturally calls for multi-class mask prediction, whose central challenge is defining the meaning of each class ID for an open-ended image and instruction.
The two-phase design provides a solution: while Binary All-Mask Prediction uses Phase~1 mainly to generate a descriptive response and the \texttt{<SEG>} trigger, our extension uses it to produce a structured target list that assigns an ID to each semantic category or object instance, optionally with a box for instance disambiguation.
Conditioned on the Phase~1 structured list, Phase~2 uses a multi-class head whose foreground classes are aligned one-to-one with the generated IDs.
In a single forward pass, each image-aligned mask token is assigned either to the background or to an ID---and thus to the corresponding semantic category or object instance defined in Phase~1---yielding a unified structured mask map.
We term this extension \emph{Structured All-Mask Prediction} (\cf \cref{fig:comp_with_pp}b).

Under this paradigm, we develop \modelP{}, a unified MLLM segmentor that instantiates Structured All-Mask Prediction across heterogeneous tasks and spatial regimes.
Extending from binary target segmentation to broader multi-target settings introduces two requirements absent from \model{}.
First, whereas \model{} collapses each query-defined target into one foreground/background mask~\cite{liu2026STAMP}, multi-target tasks require a variable set of semantic categories or object instances to retain explicit identities in a shared prediction.
To meet this requirement, \modelP{} redesigns the Phase~1 response protocol to generate structured target definitions and the Phase~2 supervision to bind their IDs to a shared multi-class mask label space.
Second, broader task coverage is meaningful only if the same model and interface operate across tasks; separately fine-tuning a checkpoint for each benchmark would demonstrate task adaptation rather than the generality of the paradigm.
We therefore train \modelP{} on a unified mixture spanning referring, reasoning, open-vocabulary semantic, and instance-aware segmentation, using the same target-ID interface and a single checkpoint throughout.
This design preserves the binary target-segmentation ability of \model{} while adding identity-preserving multi-target prediction without task-specific fine-tuning.

Extending the task coverage also exposes a complementary spatial challenge: small-target scenes require substantially finer sampling than ordinary natural images.
\model{} is designed primarily for ordinary natural-image resolutions and typically uses at most 1280 image-aligned mask tokens, leaving limited mask resolution for targets that occupy only a small fraction of the image.
To extend the same All-Mask interface to this regime, \modelP{} selectively increases the input resolution and the aligned mask-token budget, using up to 3200 tokens when finer spatial sampling is needed. This high-resolution scaling preserves a denser patch--token grid for small targets while leaving the structured prediction mechanism unchanged; its effectiveness is validated through controlled comparisons on remote-sensing benchmarks.
Our experiments assess both the resulting capabilities and the mechanisms underlying them.
Evaluations spanning the unified task suite and remote-sensing small-target segmentation show that \modelP{} combines strong segmentation accuracy, preserved general dialogue ability, and efficient single-pass mask prediction, thereby resolving the trilemma beyond the original binary target setting.
Complementary analyses examine how structured target generation in Phase~1 improves Phase~2 mask prediction and whether the spatial grounding learned through segmentation can, in turn, enhance the model's own visual understanding.

Overall, \modelP{} not only extends \model{} across task settings, output structures, and spatial regimes, but also provides diagnostic analyses of the two-phase interaction and reveals the potential of segmentation-induced grounding to further improve multimodal understanding. Our key contributions are summarized as follows:

\noindent In \textbf{\model}:
\begin{itemize}

    \item \textbf{The Trilemma Resolved for Task-Specific Single-Target Segmentation.} We instantiate the paradigm as Binary All-Mask Prediction and develop \model{}, which is fine-tuned separately for different single-target segmentation tasks. Across these tasks, \model{} achieves strong segmentation performance and efficient mask generation; when jointly trained with visual-instruction data, it also preserves the general dialogue ability of its backbone.
\end{itemize}

\noindent In \textbf{\modelP}:
\begin{itemize}
    \item \textbf{The Trilemma Resolved for Broader Segmentation Settings.} \modelP{} newly supports remote-sensing small-target segmentation, open-vocabulary semantic segmentation, and instance-aware segmentation. Across these evaluated settings, it maintains strong segmentation performance and general dialogue ability while retaining efficient single-pass mask prediction.

    \item \textbf{One Unified Model for All Evaluated Segmentation Tasks.} We develop \modelP{} as a single checkpoint covering referring, reasoning, open-vocabulary semantic, instance-aware, and remote-sensing small-target segmentation without task-specific fine-tuning.

    \item \textbf{Diagnostic Analysis and Understanding Potential.} Beyond expanding task coverage, we analyze how structured Phase~1 generation improves Phase~2 segmentation and how segmentation-learned spatial grounding can support the model's own visual understanding.
\end{itemize}

\noindent\textbf{Codebase.}
The complete codebase, including data preparation, training, inference, and evaluation scripts together with detailed documentation, is included in the supplementary material.
\section{Related Work}
\label{sec:rel}

\subsection{Multimodal Large Language Models (MLLMs)} 

The advent of MLLMs, powered by the advanced reasoning and instruction-following abilities of their LLM foundations~\citep{kaplan2020scaling, openai2024hello, gemini, liu2025empowering}, has established a new frontier for instruction-driven vision tasks. 
Most existing MLLMs are built by connecting a visual encoder with a pre-trained LLM through lightweight modality adapters, query transformers, or cross-attention modules. 
Representative architectures such as Flamingo~\citep{alayrac2022flamingo},  InstructBLIP~\citep{dai2023instructblip}, LLaVA~\citep{llava, liu2024improved, li2024llava}, and Qwen-VL~\citep{bai2023qwenvl, bai2025qwen2, yang2025qwen3} enable language models to interpret visual inputs and follow complex multimodal instructions. 
Recent models further enhance visual perception through high-resolution inputs~\cite{zhu2025internvl3}, dynamic-resolution processing~\cite{bai2025qwen3vl}, and improved visual token representations~\cite{li2025latent}, leading to strong performance on general visual question answering, OCR, document understanding, chart understanding, and visual grounding. 
Despite these advances, most MLLMs still primarily produce discrete textual responses, making it non-trivial to extend them to dense prediction tasks such as image segmentation, where the desired output is a pixel-level or patch-level mask rather than a short natural-language answer.

The central challenge for MLLM-based segmentation is therefore how to translate the high-level, spatially aware understanding of MLLMs into accurate dense masks while preserving their general dialogue ability and efficient inference~\cite{liu2026STAMP}. 
Existing methods mainly follow two strategies. 
One line of work treats the MLLM as a powerful vision-language encoder and predicts task-specific embeddings to control an external mask decoder~\citep{lai2024lisa, xia2024gsva, ren2024pixellm, rasheed2024glamm, read}. 
While effective for mask quality, such decoder-oriented supervision may interfere with the MLLM's language modeling space and weaken its general conversational ability. 
Another line of work preserves the native language-modeling interface by reformulating segmentation as next-token prediction, where masks are represented through coordinates, polygons, chain-of-thought descriptions, or patch-wise textual labels~\citep{lan2024text4seg, wu2024visionllm, liu2025seg, zhu2025segagent}. 
However, sparse textual representations often suffer from error accumulation, whereas richer textual masks require long autoregressive sequences and thus lead to slow inference. 
These limitations motivate a segmentation paradigm that can stay within the MLLM token interface while avoiding autoregressive dense mask generation.

\subsection{MLLM-Based Segmentation} Existing MLLM-based segmentation methods can be categorized into two primary paradigms: \emph{Embedding Prediction} and \emph{Next-Token Prediction} (\cf \cref{tab:comp-paradigm}). They are distinguished by the MLLM's output format for mask generation. The former employs the MLLM to produce continuous embeddings that guide an external mask decoder. The latter, however, uses the MLLM to generate a sequence of discrete tokens which, in turn, define the segmentation mask.

\noindent{\textbf{Embedding Prediction.}} 
This paradigm, pioneered by LISA~\citep{lai2024lisa}, trains an MLLM to output a special token (\eg, \texttt{[SEG]}) whose embedding prompts an external, SAM-like decoder~\citep{kirillov2023segment}. In this design, the MLLM mainly localizes the referred target at the semantic level, while the external decoder is responsible for producing the dense mask. 
As noted in \cref{tab:comp-paradigm}, this approach, refined by subsequent works~\cite{ren2024pixellm, xia2024gsva, jang2025mmr, read, wang2025segllm}, yields an efficient $\mathcal{O}(1)$ mask generation step (for one target), since the MLLM only needs to generate a single special token to initiate mask decoding. However, this efficiency is tied to a binary target-mask formulation: each special token typically prompts the decoder to produce one target/background mask. When distinct objects or instances are required, these methods usually rely on generating multiple target-specific embeddings or repeatedly invoking the same binary decoding process, rather than producing a unified semantic or instance-level mask map in one shared label space. In addition, the mandatory external module means these methods are not mask decoder-free and require structural modifications to the MLLM architecture. Supervising the MLLM-to-decoder pathway with pixel-level mask losses departs from token-only supervision, forcing the MLLM hidden states to adapt to the decoder's feature space and potentially compromising general dialogue abilities.

\noindent{\textbf{Next-Token Prediction.}} This paradigm avoids objective conflicts by representing the mask as a sequence of discrete tokens, keeping segmentation within the native language-modeling interface. Early works like VisionLLM~\citep{wang2023visionllm} generate sparse polygon coordinates, resulting in a decoder-free design, but such representations are fragile since one inaccurate coordinate or ordering error can corrupt the entire mask. To improve robustness, methods such as Seg-Zero~\citep{liu2025seg} and SegAgent~\citep{zhu2025segagent} use CoT reasoning to plan keypoints before calling an external SAM decoder. A different strategy, proposed by Text4Seg~\citep{lan2024text4seg}, autoregressively generates patch-wise classifications to remain decoder-free and provide denser mask descriptions. Text4Seg++ further reduces the per-target generation length by first predicting a bounding box and then generating foreground/background labels only within the cropped box region. However, these improvements still remain within a target-specific binary prediction formulation: for each referred object or instance, the model must generate an independent sequence of mask tokens and produce one target/background mask at a time. 
Therefore, although Text4Seg++ alleviates the autoregressive burden for one binary target mask, its inference cost still scales with the number of distinct targets in multi-category or instance-aware scenarios.
As shown in \cref{tab:comp-paradigm}, the number of generation steps still scales with both the per-target output length and the number of targets, making high-quality multi-target segmentation prohibitively slow.

\subsection{Perception for Multimodal Understanding}
Beyond producing perceptual outputs as final predictions, recent studies have shown that preliminary perception can also serve as useful intermediate evidence for multimodal understanding~\cite{bai2025qwen3vl, zhu2025internvl3, zhang2025mllms, wang2025mllm}. One line of work integrates discrete perceptual supervision directly into MLLMs, such as grounding text spans with bounding boxes or generating spatial coordinates, enabling models to localize visual entities and reason over them in language~\citep{peng2023kosmos, chen2023shikra, bigverdi2025perception}. 
These methods suggest that explicit spatial localization can provide useful visual anchors for downstream reasoning, but their perceptual signals are usually sparse, relying mainly on points, boxes tokens rather than dense object- or instance-level masks. 
Another line of work augments MLLMs with external perception tools, where the model calls detectors, segmentors, OCR models, or visual editing tools to revisit the image and obtain additional visual evidence before answering~\citep{suris2023vipergpt, lu2023chameleon, yang2023mmreact, hu2024visualsketchpad}. 
While effective, these approaches depend on external tool pipelines and do not endow the MLLM itself with native dense perception capabilities. 
More recently, latent visual reasoning methods explore performing intermediate reasoning in continuous or visual latent spaces, for example by introducing perception tokens, latent visual tokens, or continuous multimodal thought states~\citep{bigverdi2025perception, yang2025mirage, li2025latent, pham2025mcout}. 
These methods further support the view that visual intermediate representations can benefit reasoning, but they do not provide explicit, controllable, and interpretable segmentation maps as reusable visual context.

In contrast, \modelP treats segmentation as a native dense-perception interface for MLLMs. It natively supports joint multi-target segmentation while explicitly investigating whether the spatial grounding learned through structured segmentation can benefit downstream multimodal understanding. Beyond producing unified semantic- or instance-aware mask maps in a single pass, we examine whether segmentation-induced spatial representations can be reused in reasoning.
\section{Method}

\begin{figure*}[t]
    \centering
    \includegraphics[width=0.82\linewidth]{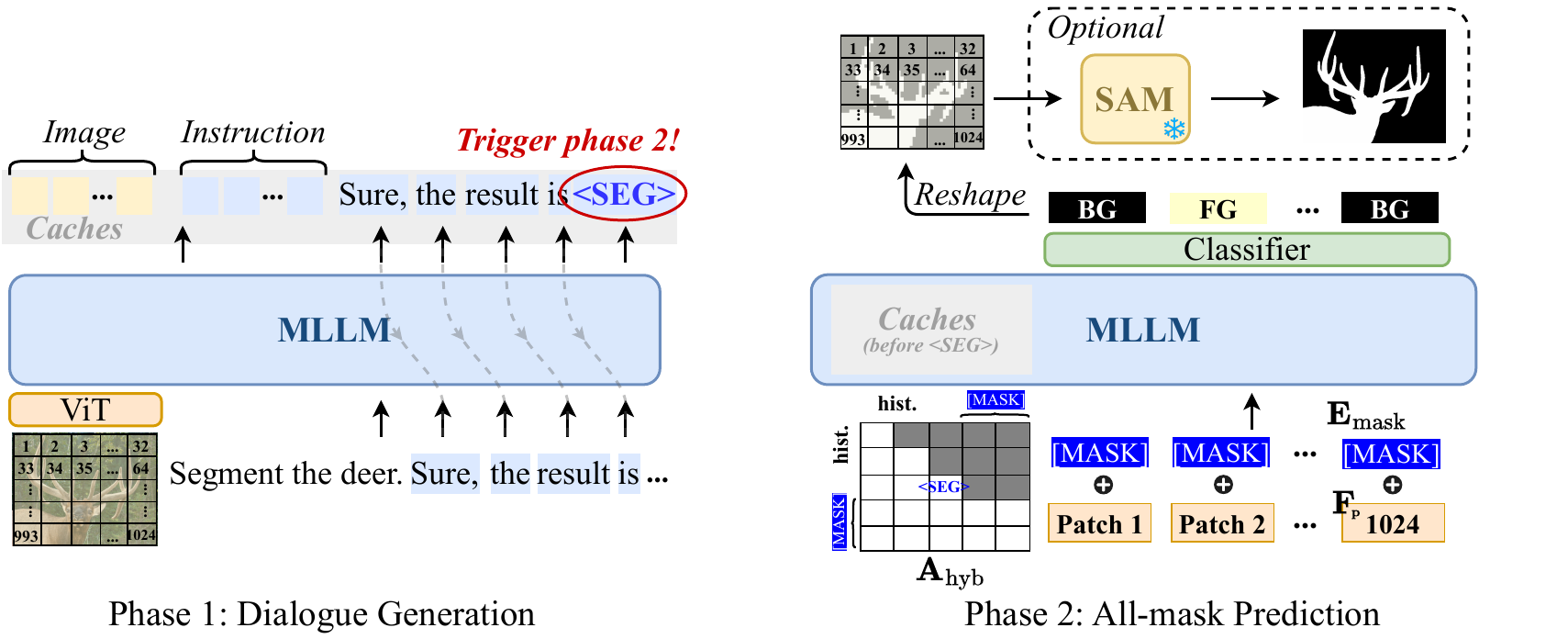}
    \caption{\textbf{The \model{} Pipeline.} 
\textbf{Phase 1 (Dialogue Generation):} The MLLM autoregressively generates a conversational response, emitting a special \texttt{<SEG>} token to trigger Phase~2.
\textbf{Phase 2 (Binary All-Mask Prediction):} Triggered by \texttt{<SEG>}, image-aligned \texttt{[MASK]} tokens are prefilled and fused with their corresponding image-patch features. A single non-autoregressive forward pass with hybrid attention predicts their foreground/background labels simultaneously; the resulting patch mask can optionally prompt a frozen SAM for refinement.
}
    \label{fig:methods}
\end{figure*}

In this section, we first detail \emph{Binary All-Mask Prediction}, instantiated by \model{} for task-specific single-target segmentation. Each query-defined target is represented by one binary foreground/background mask. As shown in \cref{fig:methods}, Phase~1 autoregressively generates a response and emits \texttt{<SEG>}, while Phase~2 predicts all image-aligned mask tokens in one non-autoregressive forward pass.

We then present \emph{Structured All-Mask Prediction}, instantiated by \modelP{} for broader multi-target settings. As shown in \cref{fig:methodsPP}, \modelP{} retains the same \texttt{<SEG>} trigger, mask-token prefilling, hybrid attention, and single-pass prediction, but introduces two structural extensions: Phase~1 generates a structured target list with explicit target IDs, and Phase~2 dynamically binds these IDs to a shared multi-class mask label space. To support spatial regimes with much smaller targets, \modelP{} further scales the input resolution together with the aligned mask-token budget, preserving a denser prediction grid without changing the structured formulation.

\newcommand{\vect}[1]{\mathbf{#1}}
\newcommand{\set}[1]{\mathcal{#1}}
\newcommand{\code}[1]{\texttt{#1}}

\subsection{\model{}: Binary All-Mask Prediction}
\label{sec:stamp_binary}

\subsubsection{Dialogue Generation (Phase~1)}
In Phase 1, the model generates a contextual textual response and identifies segmentation targets. Given an image $I$ and an instruction $T$, a ViT first extracts patch features $\mathbf{F_p} \in \mathbb{R}^{N \times D}$, which are then prepended to the text embeddings. The MLLM processes this combined input to autoregressively generate a response $R$. During this process, the model can emit a special \texttt{<SEG>} token from its vocabulary. Critically, unlike methods such as LISA~\citep{lai2024lisa} that tie the \texttt{[SEG]} token's embedding to an external decoder, our \texttt{<SEG>} token is a standard vocabulary item. It functions purely as a learned, in-vocabulary signal that triggers Phase 2, as depicted in~\cref{fig:methods} (left).

\textbf{Caches $\set{C}$.} To ensure an efficient transition, intermediate representations from Phase 1 are cached. For each \texttt{<SEG>} token, we define a context-specific tuple $(\text{hist}_i, \text{cache}_i)$, where $\text{hist}_i$ is the dialogue history leading to the token, and $\text{cache}_i$ stores the pre-computed key-value (KV) states for all preceding tokens. Preserving these states allows \model{} to avoid redundant computation when initiating mask prediction for each target.

\subsubsection{All-Mask Prediction (Phase~2)}
\label{sec:phase2}

Upon the emission of a \texttt{<SEG>} token, Phase 2 generates the entire mask in a single, non-autoregressive forward pass. This begins by preparing a specialized input sequence $\mathbf{S}_\text{in}$. We take the dialogue history preceding the \texttt{<SEG>} token and append $N$ \texttt{[MASK]} placeholders, one for each image patch. To provide spatial context, the initial embedding of each \texttt{[MASK]} token is fused with its corresponding patch feature from $\mathbf{F_p}$, along with the positional embedding that specifies the patch's location in the original image grid. We term these visually-augmented mask embeddings $\mathbf{E}_{\text{mask}}$.

A key component of this phase is our hybrid attention mechanism, which uses a custom attention mask $\mathbf{A}_\text{hyb}$ to partition the sequence. For the dialogue history, it enforces standard causal attention. For the $\mathbf{E}_{\text{mask}}$, it enables bi-directional attention, allowing each placeholder to attend to the entire context and all other placeholders. This ensures a holistic and context-aware prediction, as shown in~\cref{fig:methods}.

With this setup, the MLLM performs a single forward pass. The KV states for the dialogue history are efficiently loaded from the Phase 1 cache. The final hidden states $Z_\text{mask}$ corresponding to the \texttt{[MASK]} input are then fed into a linear classifier to produce foreground (FG) or background (BG) logits for each patch. Finally, this patch-level output can be optionally refined into a high-resolution mask by sampling several keypoints from the prediction to prompt a frozen SAM decoder~\citep{lan2024text4seg}.

\begin{figure*}[t]
    \centering
    \includegraphics[width=\linewidth]{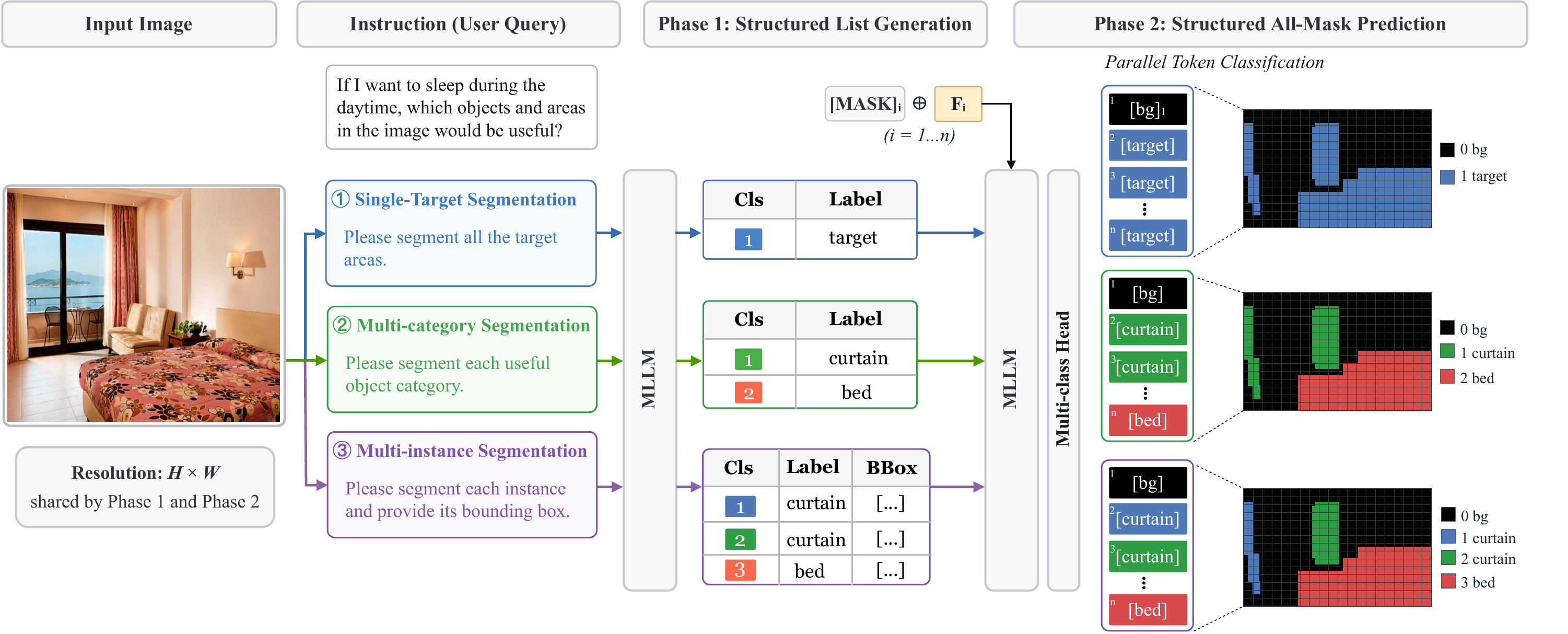}
    \caption{\textbf{The \modelP{} Pipeline.}
\textbf{Phase 1 (Structured Target Generation):}
The MLLM generates a target list at the single-target, multi-category, or multi-instance granularity requested by the instruction; each entry receives an ID and may include a bounding box for instance disambiguation.
\textbf{Phase 2 (Structured All-Mask Prediction):}
Image-aligned \texttt{[MASK]} tokens are classified simultaneously in one non-autoregressive forward pass, with the output classes dynamically bound to the target IDs specified by the \texttt{cls} fields generated in Phase~1.
}
    \label{fig:methodsPP}
\end{figure*}

\subsection{\modelP{}: Structured All-Mask Prediction}
\label{sec:stampp_structured}

While \model{} provides an efficient formulation for binary target segmentation, complex instructions often require multiple targets to be segmented and distinguished simultaneously.
Instead of repeatedly invoking binary mask prediction for different targets, \modelP{} generalizes the output space from a binary foreground/background mask to an instruction-defined structured label map. 
Importantly, \modelP{} inherits the core all-mask mechanism of \model{}, including the \texttt{<SEG>} trigger, mask-token prefilling, visual feature fusion, KV-cache reuse, hybrid attention, and single-pass non-autoregressive prediction. 
Therefore, we focus on the extensions introduced by \modelP{}: structured target representation in Phase~1 and instruction-defined multi-class mask prediction in Phase~2.

\subsubsection{Structured Target Generation (Phase~1)}
In Phase~1, \modelP{} follows the same generation logic as \model{}: the MLLM autoregressively produces a textual response and emits \texttt{<SEG>} as an in-vocabulary trigger for Phase~2. 
The difference lies in what is generated before this trigger. 
In the original \model{}, the generated response mainly serves as a free-form textual specification of a segmentation target. 
While this is sufficient for binary target segmentation, it does not provide an explicit and parseable mapping between multiple target descriptions and the mask IDs predicted in Phase~2.
For more complex scenarios, we expect Phase~1 to enumerate all segmentation targets in advance and establish a clear correspondence between each target and its \texttt{cls} value in the structured mask map predicted by Phase~2.

To this end, \modelP{} is trained to generate a \emph{JSON-style} structured target list before \texttt{<SEG>}. 
This format is compact, machine-readable, and naturally supports instruction-controlled target specification. 
Each entry follows the basic format \texttt{\{"cls": ID, "label": object\}}, which builds an explicit mapping from a numerical mask ID to a textual target name. 
The \texttt{cls} field denotes the sample-specific class ID assigned to a target: class 0 is reserved for background, while positive IDs identify the generated targets and are dynamically bound to the Phase~2 output classes. 
When required by the instruction, we further include auxiliary fields such as \texttt{"bbox"} to provide spatial cues for instance disambiguation.

As illustrated in \cref{fig:methodsPP}, different instructions can request different segmentation granularities, and \modelP{} responds by generating the corresponding structured target list. 
For single-target segmentation, the list uses a fixed \texttt{target} label to denote all regions satisfying the instruction, without distinguishing object categories or instances. 
For multi-category segmentation, the list assigns different \texttt{cls} values to different target object categories or object types, enabling the output mask map to separate multiple requested objects. 
For multi-instance segmentation, the list further assigns different \texttt{cls} values to individual instances. 
In this case, spatial fields such as \texttt{"bbox"} are important for disambiguation, since two instances may share the same textual name, \eg, two curtains in the same image, but should correspond to different mask IDs.

Thus, \modelP{} preserves the autoregressive generation and \texttt{<SEG>} triggering mechanism of \model{}, but changes the generated content from a free-form binary target specification into an instruction-conditioned, parseable ID-to-target mapping.
This design allows the same model to support broader segmentation within a unified framework, while dynamically defining the label space used by Phase~2.

\subsubsection{Structured All-Mask Prediction (Phase~2)}
In Phase~2, \modelP{} preserves the core all-mask mechanism of \model{}, including mask-token prefilling, visual feature fusion, KV-cache reuse, hybrid attention, and single-pass non-autoregressive prediction. 
The difference lies in the prediction head and its dynamic label binding. 
Instead of the binary foreground/background classifier used in \model{}, \modelP{} adopts a multi-class mask head with a fixed class capacity, set to 200 in our implementation. Samples containing more than 199 target entries are excluded or truncated during preprocessing; no evaluated sample exceeds this capacity.

Importantly, this class space is not a fixed semantic taxonomy. 
Since Phase~2 reuses the dialogue history and KV cache from Phase~1, the structured target list generated before \texttt{<SEG>} remains part of the prediction context. 
Therefore, the MLLM can condition the mask-token representations on the generated ID-to-target mapping. 
For example, in the instance-level case shown in \cref{fig:methodsPP}, Phase~1 may generate three targets, \eg, \texttt{\{"cls": 1, "label": "curtain", "bbox": ...\}}. 
Conditioned on this structured list, the multi-class mask head predicts over the fixed class space, where class 0 denotes background, class 1 is bound to the first curtain, class 2 to the second curtain, and class 3 to the bed for this input. 
The remaining numerical IDs are still part of the head capacity, but they are not assigned target semantics by the current Phase~1 list. 
Accordingly, the Phase~1 target list defines a sample-specific active ID set $\mathcal{K}$. Predictions outside $\{0\}\cup\mathcal{K}$ are treated as invalid and excluded from the final structured mask map; such predictions occur rarely in practice.
Thus, the same numerical ID can refer to different targets across different instructions, but within each sample it is dynamically bound to the target specified in Phase~1.
After prediction, mask tokens assigned the same ID are grouped as the region of the corresponding target, producing a structured mask map aligned with the Phase~1 target list.

Compared with \model{}, \modelP{} makes the interaction between the two phases more explicit. 
In \model{}, Phase~1 mainly serves as a trigger and textual context for binary mask prediction. 
In \modelP{}, Phase~1 additionally defines an instruction-specific label space through a structured ID-to-target mapping, which is then used by Phase~2 to produce the structured mask map. 
Therefore, the generated target list is not merely an auxiliary textual output, but an important context that guides mask prediction. 
This design also opens a natural way to improve structured segmentation: enhancing the quality of Phase~1 target generation can directly benefit the Phase~2 mask prediction, as further validated in our experiments.

\subsubsection{High-Resolution Mask-Token Scaling}
\label{sec:adaptive_mask_tokens}

The image-aligned design of All-Mask Prediction allows its spatial granularity to vary with the input resolution. Let $N$ denote both the number of retained image-patch features and the number of corresponding \texttt{[MASK]} placeholders. Resizing every image to a fixed token grid is adequate for common natural images, but it can be harmful in small-target scenes, where the target occupies only a small image fraction. In particular, aggressive downsampling may leave a small target represented by very few tokens or remove it entirely from the retained visual evidence.

We exploit this image-aligned property through high-resolution mask-token scaling. For ordinary natural images, we use $N=1024$--$1280$ image-aligned mask tokens. For remote-sensing small-target inputs, we retain a higher input resolution and correspondingly increase the budget to $N=2560$--$3200$. Scaling the two together is important: the higher resolution retains more patch-level evidence, while the larger aligned token budget preserves this finer sampling in the predicted mask grid. This extension changes only the input and mask-token sequence lengths; the structured ID binding, hybrid attention, and multi-class head remain unchanged, and all mask tokens are still predicted together in one non-autoregressive forward pass. Its effect is isolated through the controlled study in \cref{sec:exp_ablation}.

\subsection{Training}
We train \model{} and \modelP{} with a unified objective consisting of a text generation loss $\mathcal{L}_{\text{text}}$ and a mask prediction loss $\mathcal{L}_{\text{mask}}$:
\begin{equation}
    \mathcal{L} = \mathcal{L}_{\text{text}} + \mathcal{L}_{\text{mask}}.
\end{equation}

The text loss $\mathcal{L}_{\text{text}}$ is the standard cross-entropy loss for autoregressive language modeling. 
For a ground-truth response $Y = (y_1, \dots, y_L)$, it is defined as:
\begin{equation}
    \mathcal{L}_{\text{text}} = - \sum_{i=1}^{L} \log P(y_i | y_{<i}, I, T).
\end{equation}
For \model{}, the target response contains the textual answer and the \texttt{<SEG>} trigger. 
For \modelP{}, we do not introduce additional supervision terms for structured target generation; instead, the target response in Phase~1 is formatted as a JSON-style structured target list before \texttt{<SEG>}. 
Thus, the same autoregressive text loss naturally trains the model to generate parseable structured outputs and establish the instruction-specific ID-to-target mapping used by Phase~2.

The mask loss $\mathcal{L}_{\text{mask}}$ is applied to the logits of the mask tokens. 
Let $N$ denote the number of mask tokens. 
For \model{}, each mask token performs binary foreground/background classification. 
Given the predicted foreground probability $p_j$ and binary ground-truth label $y_j \in \{0,1\}$ for the $j$-th mask token, we use Binary Cross-Entropy and Dice losses:
\begin{equation}
    \mathcal{L}_{\text{BCE}}
    =
    -\frac{1}{N}\sum_{j=1}^{N}
    \left[
    y_j \log p_j + (1-y_j)\log(1-p_j)
    \right],
\end{equation}
\begin{equation}
    \mathcal{L}_{\text{Dice}}
    =
    1 -
    \frac{2\sum_{j=1}^{N} p_j y_j + \epsilon}
    {\sum_{j=1}^{N} p_j + \sum_{j=1}^{N} y_j + \epsilon}.
\end{equation}
The binary mask loss is:
\begin{equation}
\mathcal{L}_{\text{mask}}^{\mathrm{STAMP}}
=
\mathcal{L}_{\text{BCE}}
+
\mathcal{L}_{\text{Dice}}.
\end{equation}

For \modelP{}, the ground-truth masks are converted into a structured patch-level label map according to the target IDs specified in the ground-truth Phase~1 response. During inference, the label mapping is instead determined by the structured target list generated by the model. 
Each mask token is supervised to predict either background or one of the instruction-defined foreground IDs. 
Let $C$ denote the fixed class capacity of the multi-class mask head, and let $\mathbf{p}_{j} \in \mathbb{R}^{C}$ be the predicted class distribution for the $j$-th mask token. 
The one-hot ground-truth label is denoted as $\mathbf{y}_{j} \in \{0,1\}^{C}$. 
We use multi-class cross-entropy:
\begin{equation}
    \mathcal{L}_{\text{CE}}
    =
    -\frac{1}{N}\sum_{j=1}^{N}
    \sum_{c=0}^{C-1}
    y_{j,c} \log p_{j,c}.
\end{equation}

To preserve region-level mask quality, we further apply a class-wise Dice loss over the segmentation channels:
\begin{equation}
    \mathcal{L}_{\text{Dice}}
    =
    1 -
    \frac{1}{C}
    \sum_{c=0}^{C-1}
    \frac{
    2\sum_{j=1}^{N} p_{j,c} y_{j,c} + \epsilon
    }{
    \sum_{j=1}^{N} p_{j,c}
    +
    \sum_{j=1}^{N} y_{j,c}
    +
    \epsilon
    },
\end{equation}
where \(C\) denotes the total number of segmentation channels, including the background channel.

The structured mask loss is:
\begin{equation}
    \mathcal{L}_{\text{mask}}^{\mathrm{STAMPlus}}
    =
    \mathcal{L}_{\text{CE}} + \mathcal{L}_{\text{Dice}}.
\end{equation}

In this sense, \model{} can be viewed as a binary special case of \modelP{}, where the structured label space contains only background and one foreground target. 
By jointly optimizing text generation and mask-token classification, the model learns not only when to invoke mask prediction, but also how to align the structured target specification generated in Phase~1 with the structured mask map predicted in Phase~2.
\section{Experiments}
\label{sec:experiments}

We organize our experiments by task to evaluate \model{} and \modelP{} progressively:

\begin{itemize}
    \item \textbf{Single-Target Referring and Reasoning Segmentation.}
    We first evaluate \model{} and test whether the unified \modelP{} retains its established single-target segmentation ability on natural-image benchmarks. We then extend the evaluation to remote-sensing images with small targets (\cref{sec:exp_binary_target}).

    \item \textbf{Open-Vocabulary Semantic and Instance-Aware Segmentation.}
    We evaluate the new ability of \modelP{} to distinguish multiple semantic categories or object instances in a single prediction (\cref{sec:exp_structured_seg}).

    \item \textbf{Segmentation--Understanding Interaction.}
    We study whether better Phase~1 target descriptions improve Phase~2 segmentation and whether the spatial grounding learned through segmentation can support downstream visual understanding (\cref{sec:exp_seg_understanding}).

    \item \textbf{General Ability, Efficiency, and Ablations.}
    We evaluate general multimodal ability, inference efficiency, and the main design choices and generalization properties of All-Mask Prediction (\cref{sec:exp_general_efficiency,sec:exp_ablation}).
\end{itemize}

This evaluation directly mirrors the trilemma posed in \cref{sec:intro}: the task-specific experiments measure segmentation performance across broader settings, the general multimodal benchmarks assess dialogue compatibility, and the latency studies test whether the expanded capabilities retain efficient inference.

\subsection{Common Setup}
\label{sec:exp_setup}

\noindent\textbf{Implementation Details.}
We build 2B and 7B variants of \model{} and \modelP{} mainly on Qwen2-VL~\cite{wang2024qwen2}, with an additional LLaVA-based variant for backbone-transfer analysis.
Both models share the mask-token design, hybrid attention, optional SAM-H refinement, and a linear mask head: \model{} maps a $d$-dimensional mask-token state to two foreground/background classes, whereas \modelP{} maps it to 200 structured classes.
Natural-image experiments use 1024--1280 image-aligned \texttt{[MASK]} tokens; remote-sensing experiments jointly increase the input resolution and token budget to 2560--3200 to preserve small-target evidence (\cref{sec:adaptive_mask_tokens}).

\noindent\textbf{Training.}
All models are trained on NVIDIA H800 GPUs using AdamW.
\modelP{} follows the same optimization recipe as \model{}, differing only in its structured Phase~1 response format and multi-class Phase~2 supervision.
The \model{} checkpoints follow benchmark-specific protocols, whereas all \modelP{} results at each scale use one unified checkpoint without benchmark-specific fine-tuning unless otherwise specified.

\noindent\textbf{Training Data.}
\model{} uses the natural-image referring and reasoning subsets under task-specific protocols~\cite{liu2026STAMP}.
The unified \modelP{} checkpoint is trained on a data mixture constructed from RefCLEF~\cite{kazemzadeh2014referitgame}, RefCOCO~\cite{kazemzadeh2014referitgame}, RefCOCO+~\cite{kazemzadeh2014referitgame}, RefCOCOg~\cite{mao2016generation} and gRefCOCO~\cite{liu2023gres} for standard referring segmentation; ReasonSeg~\cite{lai2024lisa} for reasoning-intensive target discovery; RRSIS-D~\cite{liu2024rotated} and EarthReason~\cite{li2025segearth} for remote-sensing referring and reasoning; COCO-Stuff~\cite{cocostuff} for dense semantic supervision; COCO Panoptic~\cite{coco} and MUSE~\cite{ren2024pixellm} for category- and instance-aware masks; and LLaVA-665k~\cite{llava} for general visual instruction following.

\noindent\textbf{Metrics.}
Across the referring and reasoning segmentation benchmarks, we primarily report cumulative intersection-over-union (cIoU) and generalized intersection-over-union (gIoU).
cIoU measures the pixel-level overlap accumulated over the dataset, whereas gIoU averages IoU over individual samples.
Additional task-specific metrics are introduced with the corresponding experiments.
Unless otherwise specified, the best and second-best results in each comparison are highlighted in \textbf{bold} and \underline{underlined}, respectively.

\subsection{Single-Target Referring and Reasoning Segmentation}
\label{sec:exp_binary_target}

We first evaluate the tasks for which \model{} was originally developed and test whether \modelP{} retains this capability.
These benchmarks use a binary target mask: even when an expression refers to several regions, all referred regions are merged into one foreground class.
We consider both natural images and remote-sensing images with small targets, with the latter evaluating the proposed high-resolution mask-token scaling.

\subsubsection{Natural Images}

The natural-image evaluation spans standard referring segmentation on the RefCOCO family, generalized referring segmentation on gRefCOCO, and reasoning segmentation on ReasonSeg.
All datasets are evaluated using their standard protocols~\citep{lai2024lisa,lan2024text4seg}.

\noindent\textbf{RefCOCO Family.}
RefCOCO, RefCOCO+, and RefCOCOg provide the standard evaluation of language-guided target segmentation, with RefCOCO+ removing location words and RefCOCOg using longer, more descriptive expressions.
For direct comparison with prior work, \model{} is trained for three epochs on 800k samples from RefCLEF and the RefCOCO family following Text4Seg~\citep{lan2024text4seg}.
We compare with specialized segmentors, embedding-prediction MLLMs, and next-token-prediction methods, including UNINEXT-L~\citep{yan2023universal}, GSVA~\citep{xia2024gsva}, READ~\citep{read}, and Text4Seg~\citep{lan2024text4seg}.
As shown in \cref{tab:final_with_citations_sorted}, \model{}-7B achieves an average cIoU of 80.7 across the RefCOCO family, improving over all prior methods, including Text4Seg++ (78.9).
The smaller \model{}-2B reaches 79.1, and the unrefined variants remain competitive without SAM-based post-processing, supporting the effectiveness of the All-Mask representation itself.
Importantly, the unified \modelP{} checkpoints preserve this ability: their 2B and 7B variants obtain average cIoUs of 79.5 and 81.0, respectively, slightly improving over the task-specific \model{} counterparts and establishing the strongest overall result.

\begin{table*}[!t]
    \centering
    \renewcommand{\arraystretch}{1.03}
    \footnotesize
    \caption{\textbf{Referring-expression segmentation on the RefCOCO family.}
    Methods are grouped by paradigm and sorted by average performance when available. 
    Avg. is the mean cIoU over the eight reported dataset splits.
    Our models are highlighted with a gray background; the best and second-best results in each column are shown in bold and underlined, respectively.
    $^\dagger$ indicates results without SAM-based mask refinement.}
    \label{tab:final_with_citations_sorted}
    
    \definecolor{highlightgray}{gray}{0.93}
    \definecolor{citegray}{gray}{0.55}
    \newcommand{\citeinfo}[1]{{\color{citegray}\footnotesize #1}}

    \setlength{\tabcolsep}{3.6pt}
    \begin{tabular}{@{}llrrrrrrrrr@{}}
        \toprule
        \multirow{2}{*}{\textbf{Method}} & \multirow{2}{*}{\textbf{LLM}} & \multicolumn{3}{c}{\textbf{RefCOCO}} & \multicolumn{3}{c}{\textbf{RefCOCO+}} & \multicolumn{2}{c}{\textbf{RefCOCOg}} & \multirow{2}{*}{\textbf{Avg.}} \\
        \cmidrule(lr){3-5} \cmidrule(lr){6-8} \cmidrule(lr){9-10}
        & & val & testA & testB & val & testA & testB & val(U) & test(U) & \\
        \midrule

        \multicolumn{11}{@{}l}{\textbf{\textit{Specialized Baselines}}} \\
        \midrule
        ReLA \citeinfo{(CVPR'23)~\cite{liu2023gres}} & BERT & 73.8 & 76.5 & 70.2 & 66.0 & 71.0 & 57.7 & 65.0 & 66.0 & 68.3 \\
        PolyFormer-L \citeinfo{(CVPR'23)~\cite{liu2023polyformer}} & BERT & 76.0 & 78.3 & 73.3 & 69.3 & 74.6 & 61.9 & 69.2 & 70.2 & 71.6 \\
        UNINEXT-L \citeinfo{(CVPR'23)~\cite{yan2023universal}} & BERT & 80.3 & 82.6 & 77.8 & 70.0 & 74.9 & 62.6 & 73.4 & 73.7 & 74.4 \\
        [3pt]

        \multicolumn{11}{@{}l}{\textbf{\textit{Paradigm: Embedding Prediction}}} \\
        \midrule
        PixelLM \citeinfo{(CVPR'24)~\cite{ren2024pixellm}} & Vicuna-7B & 73.0 & 76.5 & 68.2 & 66.3 & 71.7 & 58.3 & 69.3 & 70.5 & 69.2 \\
        LISA \citeinfo{(CVPR'24)~\cite{lai2024lisa}} & Vicuna-7B & 74.9 & 79.1 & 72.3 & 65.1 & 70.8 & 58.1 & 67.9 & 70.6 & 69.9 \\
        GSVA \citeinfo{(CVPR'24)~\cite{xia2024gsva}} & Vicuna-7B & 77.2 & 78.9 & 73.5 & 65.9 & 69.6 & 59.8 & 72.7 & 73.3 & 71.4 \\
        READ \citeinfo{(CVPR'25)~\cite{read}} & Vicuna-7B & 78.1 & 80.2 & 73.2 & 68.4 & 73.7 & 60.4 & 70.1 & 71.4 & 71.9 \\
        GSVA \citeinfo{(CVPR'24)~\cite{xia2024gsva}} & Vicuna-13B & 78.2 & 80.4 & 74.2 & 67.4 & 71.5 & 60.9 & 74.2 & 75.6 & 72.8 \\
        [3pt]

        \multicolumn{11}{@{}l}{\textbf{\textit{Paradigm: Token Prediction}}} \\
        \midrule
        Text4Seg$^\dagger$ \citeinfo{(ICLR'25)~\cite{lan2024text4seg}} & Vicuna-13B & 74.1 & 76.4 & 72.4 & 68.5 & 72.8 & 63.6 & 69.1 & 70.1 & 70.9 \\
        Text4Seg$^\dagger$ \citeinfo{(ICLR'25)~\cite{lan2024text4seg}} & InternLM2.5-7B & 74.7 & 77.4 & 71.6 & 68.5 & 73.6 & 62.9 & 70.7 & 71.6 & 71.4 \\
        Seg-Zero \citeinfo{(arXiv'25)~\cite{liu2025seg}} & Qwen2.5-3B & - & 79.3 & - & - & 73.7 & - & - & 71.5 & - \\
        Seg-Zero \citeinfo{(arXiv'25)~\cite{liu2025seg}} & Qwen2.5-7B & - & 80.3 & - & - & 76.2 & - & - & 72.6 & - \\
        SegLLM \citeinfo{(ICLR'25)~\cite{wang2025segllm}} & Vicuna-7B & 80.2 & 81.5 & 75.4 & 70.3 & 73.0 & 62.5 & 72.6 & 73.6 & 73.6 \\
        Text4Seg \citeinfo{(ICLR'25)~\cite{lan2024text4seg}} & Vicuna-7B & 79.3 & 81.9 & 76.2 & 72.1 & 77.6 & 66.1 & 72.1 & 73.9 & 74.9 \\
        \rowcolor{highlightgray}
        \model$^\dagger$ & Qwen2-2B & 77.7 & 79.4 & 76.1 & 73.4 & 76.4 & 69.7 & 74.9 & 75.1 & 75.3 \\
        Text4Seg \citeinfo{(ICLR'25)~\cite{lan2024text4seg}} & InternLM2.5-7B & 79.2 & 81.7 & 75.6 & 72.8 & 77.9 & 66.5 & 74.0 & 75.3 & 75.4 \\
        SegAgent \citeinfo{(CVPR'25)~\cite{zhu2025segagent}} & Qwen-7B & 79.7 & 81.4 & 76.6 & 72.5 & 75.8 & 66.9 & 75.1 & 75.2 & 75.4 \\
        \rowcolor{highlightgray}
        \model$^\dagger$ & Qwen2-7B & 78.1 & 79.2 & 76.8 & 74.7 & 77.6 & 70.9 & 75.7 & 76.2 & 76.2 \\
        Text4Seg \citeinfo{(ICLR'25)~\cite{lan2024text4seg}} & Vicuna-13B & 80.2 & 82.7 & 77.3 & 73.7 & 78.6 & 67.6 & 74.0 & 75.1 & 76.2 \\
        \rowcolor{highlightgray}
        \model & Vicuna-7B & 80.5 & 83.0 & 77.3 & 74.8 & 79.2 & 68.6 & 75.7 & 76.8 & 77.0 \\
        Text4Seg++ \citeinfo{(TPAMI'26)~\cite{text4segpp}} & Qwen2-7B & 81.6 & 84.1 & 78.9 & 76.9 & 81.7 & 70.9 & 78.2 & 78.9 & 78.9 \\
        \rowcolor{highlightgray}
        \model & Qwen2-2B & 81.9 & 83.7 & 79.5 & 77.1 & 80.5 & 72.7 & 78.5 & 78.8 & 79.1 \\
        \rowcolor{highlightgray}
        \modelP & Qwen2-2B & 82.0 & 84.1 & 79.4 & 78.3 & 80.3 & 73.1 & 79.4 & 79.6 & 79.5 \\
        \rowcolor{highlightgray}
        \model & Qwen2-7B & \underline{83.1} & \textbf{84.5} & \underline{80.8} & \underline{79.4} & \textbf{82.8} & \underline{74.6} & \underline{79.9} & \underline{80.4} & \underline{80.7} \\
        \rowcolor{highlightgray}
        \modelP & Qwen2-7B & \textbf{83.6} & \underline{84.3} & \textbf{81.6} & \textbf{79.9} & \underline{82.6} & \textbf{75.0} & \textbf{80.1} & \textbf{81.0} & \textbf{81.0} \\
        \bottomrule
    \end{tabular}
\end{table*}

\noindent\textbf{gRefCOCO.}
gRefCOCO extends referring segmentation to multi-object and no-target expressions, while still evaluating the merged union of all referred regions as one binary mask. We therefore categorize it as single-target segmentation because the output does not preserve the identities of individual referred objects.
Following the established protocol, the RefCOCO-trained \model{} checkpoint is further trained for two epochs on the 419k-sample gRefCOCO training split and compared with methods designed for generalized expressions, including LAVT~\citep{yang2022lavt}.
On the gRefCOCO test splits, \model{}-7B without refinement already surpasses the strongest fully equipped Text4Seg baseline (71.4 vs. 71.1 average score), while the refined model reaches 74.5.
We report the benchmark-specific \model{} results in the main comparison, whereas \modelP{} is trained once on the full unified mixture without rebalancing the gRefCOCO-specific no-target cases, which constitute only a small fraction of its training data.
For completeness, the supplementary material reports the \modelP{} results both overall and after excluding no-target queries.

\begin{table}[!t]
    \centering
    \footnotesize
    \renewcommand{\arraystretch}{1.05}
    \caption{\textbf{gRefCOCO under the single-target merged-mask protocol.}
    Although a query may refer to multiple objects, all referred regions are evaluated as one foreground mask.
    Avg. is computed over the four reported test scores.
    Our models are highlighted with a gray background; bold and underlined denote the best and second-best results, respectively. $^\dagger$ denotes evaluation without SAM-based refinement.}
    \label{tab:gref}

    \definecolor{highlightgray}{gray}{0.93}

    \setlength{\tabcolsep}{1.0pt}
    \begin{tabular}{@{}llrrrrr@{}}
        \toprule
        \multirow{2}{*}{\textbf{Method}} &
        \multirow{2}{*}{\textbf{LLM}} &
        \multicolumn{2}{c}{\textbf{Test A}} &
        \multicolumn{2}{c}{\textbf{Test B}} &
        \multirow{2}{*}{\textbf{Avg.}} \\
        \cmidrule(lr){3-4} \cmidrule(lr){5-6}
        & & gIoU & cIoU & gIoU & cIoU & \\
        \midrule
        LAVT & BERT & 65.9 & 65.3 & 55.8 & 55.0 & 60.5 \\
        LISA & Vicuna-7B & 66.3 & 68.5 & 58.8 & 60.6 & 63.6 \\
        ReLA & BERT & 70.0 & 69.3 & 61.0 & 59.9 & 65.1 \\
        LISA & Vicuna-13B & 68.2 & 69.7 & 61.8 & 62.2 & 65.5 \\
        GSVA & Vicuna-7B & 71.1 & 69.9 & 62.2 & 60.5 & 65.9 \\
        Text4Seg$^\dagger$ & InternLM2.5-7B & 69.4 & 70.9 & 63.1 & 64.1 & 66.9 \\
        Text4Seg$^\dagger$ & Vicuna-13B & 69.8 & 71.4 & 63.8 & 64.4 & 67.4 \\
        Text4Seg & InternLM2.5-7B & 75.1 & 73.8 & 67.3 & 66.6 & 70.7 \\
        \rowcolor{highlightgray}
        \model$^\dagger$ & Qwen2-2B & 73.6 & 73.7 & 67.5 & 68.1 & 70.7 \\
        Text4Seg & Vicuna-13B & 75.1 & 74.3 & 68.0 & 67.1 & 71.1 \\
        \rowcolor{highlightgray}
        \model$^\dagger$ & Qwen2-7B & 73.8 & 74.7 & 68.1 & 69.1 & 71.4 \\
        \rowcolor{highlightgray}
        \model & Qwen2-2B & \underline{76.5} & \underline{75.7} & \underline{70.0} & \underline{69.8} & \underline{73.0} \\
        \rowcolor{highlightgray}
        \model & Qwen2-7B & \textbf{77.6} & \textbf{77.2} & \textbf{71.4} & \textbf{71.6} & \textbf{74.5} \\
        \bottomrule
    \end{tabular}
\end{table}

\noindent\textbf{ReasonSeg.}
ReasonSeg requires the model to infer the intended target from reasoning-intensive language rather than a direct referring expression.
Following LISA~\citep{lai2024lisa}, the \model{}-2B RES checkpoint is further trained for one epoch on the RefCOCO family and the 239 ReasonSeg training images; we compare with OVSeg~\cite{ovseg} and recent MLLM-based segmentation methods.
As reported in \cref{tab:reasonseg}, \model{}-2B achieves an average score of 63.2, improving over READ (61.1) without explicit chain-of-thought mask generation.
\modelP{}-2B remains competitive at 61.8, while the unified \modelP{}-7B checkpoint reaches the best average result of 64.0.
The standard RES and ReasonSeg evaluations therefore show that Structured All-Mask Prediction retains the single-target segmentation ability of \model{} under both direct and reasoning-intensive instructions.

\begin{table}[!t]
    \centering
    \footnotesize
    \renewcommand{\arraystretch}{1.05}
    \caption{\textbf{Reasoning segmentation on ReasonSeg.}
    We report gIoU and cIoU on the validation and test splits, with Avg. computed over the four scores.
    Our models are highlighted with a gray background; bold and underlined denote the best and second-best results, respectively.}
    \label{tab:reasonseg}
    
    \definecolor{highlightgray}{gray}{0.93}
    \definecolor{citegray}{gray}{0.55}
    \newcommand{\citeinfo}[1]{{\color{citegray}\scriptsize #1}}

    \setlength{\tabcolsep}{3.3pt}
    \begin{tabular}{@{}llrrrrr@{}}
        \toprule
        \multirow{2}{*}{\textbf{Method}} & \multirow{2}{*}{\textbf{LLM}} & \multicolumn{2}{c}{\textbf{Val}} & \multicolumn{2}{c}{\textbf{Test}} & \multirow{2}{*}{\textbf{Avg.}} \\
        \cmidrule(lr){3-4} \cmidrule(lr){5-6}
        & & gIoU & cIoU & gIoU & cIoU & \\
        \midrule
        OVSeg & Vicuna-7B & 28.5 & 18.6 & 26.1 & 20.8 & 23.5 \\
        LISA & Vicuna-7B & 53.6 & 52.3 & 48.7 & 48.8 & 50.9 \\
        SegLLM & Vicuna-7B & 57.2 & 54.3 & 52.4 & 48.4 & 53.1 \\
        Text4Seg++ & Qwen2-7B & 59.1 & 49.5 & 57.1 & 52.1 & 54.5 \\
        Seg-Zero & Qwen2.5-7B & 62.6 & 62.0 & 57.5 & 52.0 & 58.5 \\
        READ & Vicuna-7B & 59.8 & \textbf{67.6} & 58.5 & 58.6 & 61.1 \\
        \rowcolor{highlightgray}
        \model  & Qwen2-2B & \underline{65.1} & \underline{63.9} & \underline{62.7} & 60.9 & \underline{63.2} \\
        \rowcolor{highlightgray}
        \modelP & Qwen2-2B & 62.9 & 61.4 & 61.8 & \underline{61.2} & 61.8 \\
        \rowcolor{highlightgray}
        \modelP & Qwen2-7B & \textbf{65.7} & 63.0 & \textbf{63.4} & \textbf{63.8} & \textbf{64.0} \\
        \bottomrule
    \end{tabular}
\end{table}

\subsubsection{Remote-Sensing Small-Target Segmentation}

Having established compatibility on natural images, we next test whether the same All-Mask interface can operate in a substantially more demanding spatial regime.
Remote sensing provides a particularly suitable testbed for this setting: its images often cover large spatial areas, while referred objects occupy only a tiny fraction of the pixels and may collapse to very few visual patches under the resizing used for ordinary natural images.
We therefore choose this domain to evaluate high-resolution mask-token scaling for small-target segmentation, rather than merely to show cross-domain transfer.
The higher input resolution retains more patch-level evidence around small objects.
Because each retained image patch is paired with one image-aligned mask token, the enlarged patch grid is accompanied by a larger mask-token budget, as described in \cref{sec:exp_setup}.
We evaluate this setting on RRSIS-D and EarthReason using their standard validation and test splits and compare with both specialized remote-sensing models and generalist MLLM-based segmentation methods.

\noindent\textbf{RRSIS-D.}
RRSIS-D evaluates referring-expression segmentation in large remote-sensing scenes, where referred objects are frequently small relative to the full image.
Following its standard protocol, we report Acc@0.5, gIoU, and cIoU on the validation and test splits.
On RRSIS-D, \modelP{}-7B obtains an average score of 76.2, compared with 74.4 for the strongest specialized baseline and 70.8 for Text4Seg++.

\noindent\textbf{EarthReason.}
EarthReason combines geospatial reasoning with pixel-level target localization, jointly testing instruction understanding and fine-grained perception in high-resolution imagery.
On EarthReason, the 2B and 7B variants achieve average scores of 72.5 and 74.0, respectively, improving over Text4Seg++ at 70.1.
Overall, these results extend single-target segmentation to scenes with substantially larger spatial scales and much smaller target-to-image ratios, supporting the applicability of \modelP{} to small-target perception.

\begin{figure*}[!t]
    \centering
    \includegraphics[width=\linewidth]{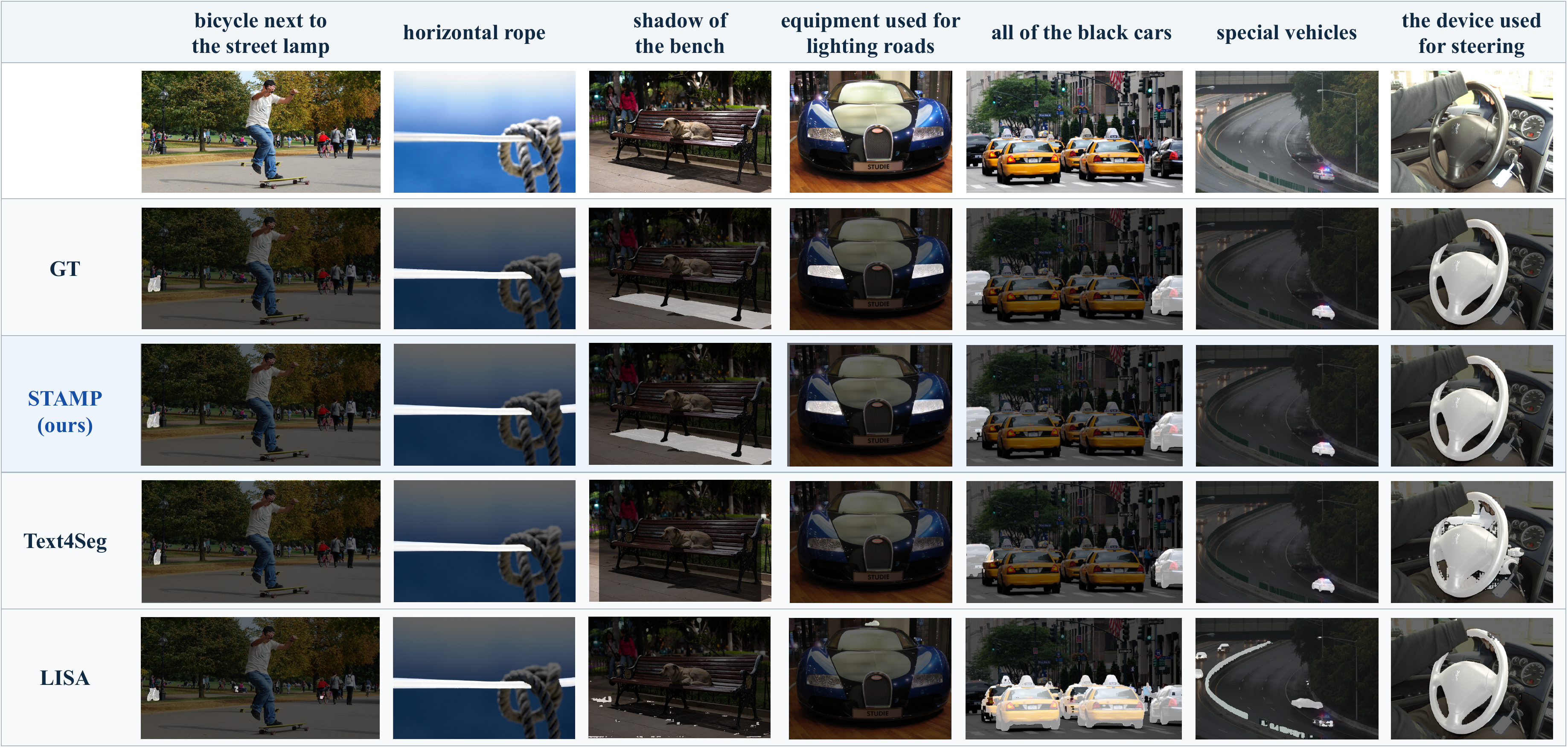}
    \caption{\textbf{Qualitative comparison on referring segmentation.}
    Each column corresponds to one query; from top to bottom, the rows show the input image, ground truth, and predictions from \model{}, Text4Seg, and LISA.}
    \label{fig:stamp_showcase}

    \vspace{0.6em}
    \includegraphics[width=\linewidth]{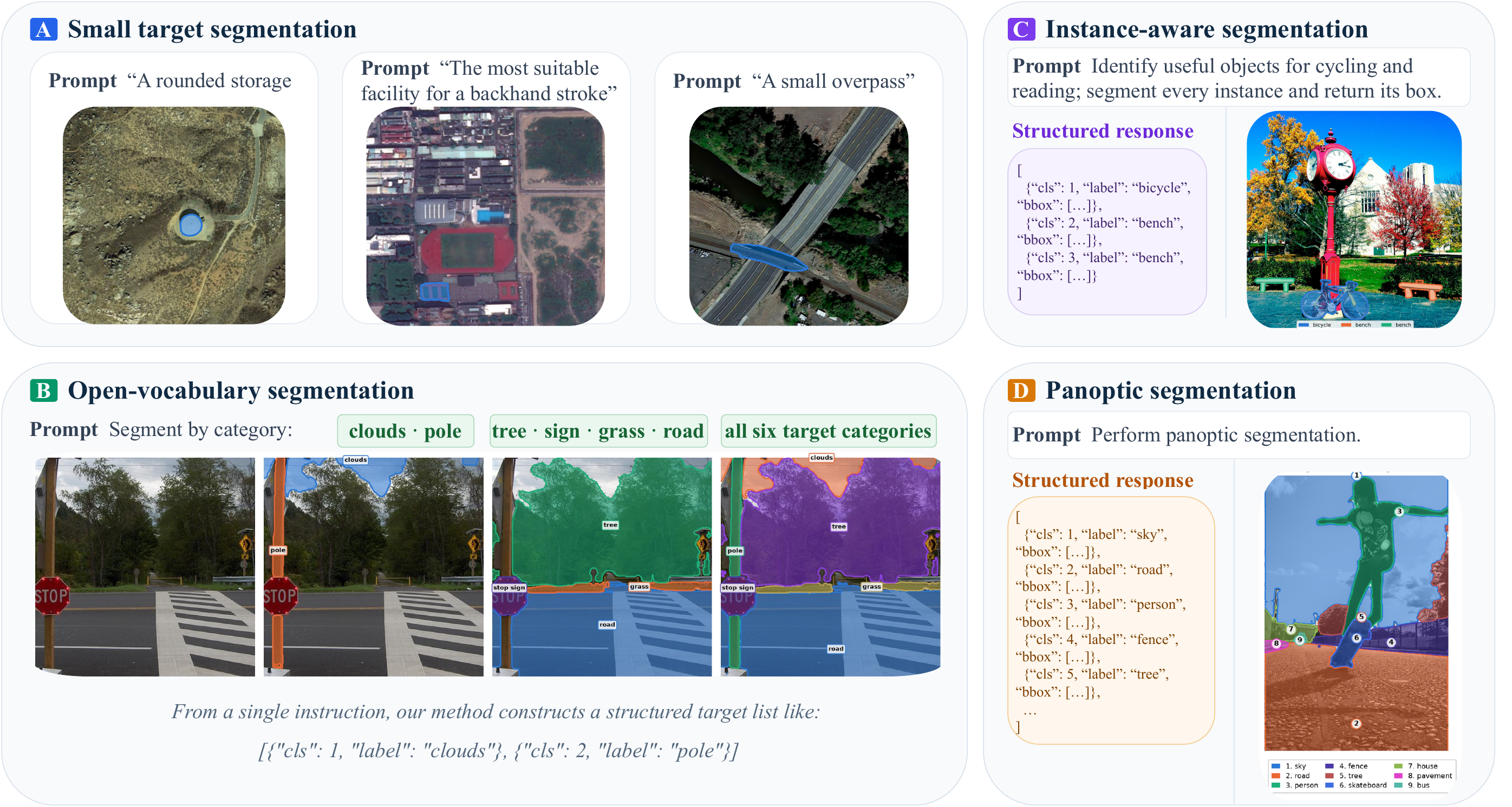}
    \caption{\textbf{Qualitative showcase of the capabilities added by \modelP{} beyond \model{}.}
    \modelP{} retains the established single-target referring and reasoning segmentation capabilities of \model{}. Because \cref{fig:stamp_showcase} already compares referring segmentation, we omit redundant single-target examples here and focus on the additional capabilities of \modelP{}:
    (a) high-resolution target segmentation for small objects in remote-sensing imagery,
    (b) open-vocabulary semantic segmentation with multiple categories predicted jointly,
    (c) instance-aware multi-target segmentation using structured target IDs and bounding boxes, and
    (d) a qualitative panoptic-style prediction with identity-preserving region labels.}
    \label{fig:stamplus_showcase}
\end{figure*}

\begin{table*}[!t]
    \centering
    \renewcommand{\arraystretch}{1.04}
    \footnotesize
    \caption{\textbf{Referring expression segmentation results on the RRSIS-D benchmark.}
    Methods are grouped by model type.
    Avg. is computed over the six validation/test metrics.
    Our models are highlighted with a gray background; the best and second-best results in each column are shown in bold and underlined, respectively.}
    \label{tab:rrsis_d_results}

    \definecolor{highlightgray}{gray}{0.93}
    \definecolor{citegray}{gray}{0.55}
    \newcommand{\citeinfo}[1]{{\color{citegray}\footnotesize #1}}

    \setlength{\tabcolsep}{6.5pt}
    \begin{tabular}{@{}llrrrrrrr@{}}
        \toprule
        \multirow{2}{*}{\textbf{Method}} &
        \multirow{2}{*}{\textbf{LLM}} &
        \multicolumn{3}{c}{\textbf{Validation Set}} &
        \multicolumn{3}{c}{\textbf{Test Set}} &
        \multirow{2}{*}{\textbf{Avg.}} \\
        \cmidrule(lr){3-5} \cmidrule(lr){6-8}
        & & Acc@0.5 & gIoU & cIoU & Acc@0.5 & gIoU & cIoU & \\
        \midrule

        \multicolumn{9}{@{}l}{\textbf{\textit{Specialized Baselines}}} \\
        \midrule
        RMSIN \citeinfo{(CVPR'24)~\cite{liu2024rotated}}
        & BERT & 74.7 & 65.1 & 78.3 & 74.3 & 64.2 & 77.8 & 72.4 \\
        LAVT \citeinfo{(TPAMI'25)~\cite{yang2022lavt}}
        & BERT & 69.5 & 61.5 & 77.6 & 69.5 & 61.0 & 77.2 & 69.4 \\
        SegEarth-R1 (FT) \citeinfo{(arXiv'25)~\cite{li2025segearth}}
        & Phi-1.5-1.3B & \textbf{78.6} & \underline{67.6} & 78.9 & \textbf{77.0} & \underline{66.4} & 78.0 & \underline{74.4} \\
        [3pt]

        \multicolumn{9}{@{}l}{\textbf{\textit{Generalist Models}}} \\
        \midrule
        LISA \citeinfo{(CVPR'24)~\cite{lai2024lisa}}
        & Vicuna-7B & 27.1 & 27.8 & - & 24.5 & 26.8 & - & - \\
        PixelLM \citeinfo{(CVPR'24)~\cite{ren2024pixellm}}
        & Vicuna-7B & 33.5 & 33.7 & - & 28.8 & 31.7 & - & - \\
        NExT-Chat \citeinfo{(ICML'24)~\cite{nextchat}}
        & Vicuna-7B & 29.0 & 27.0 & - & 26.4 & 25.0 & - & - \\
        GeoGround \citeinfo{(arXiv'24)~\cite{geoground}}
        & Vicuna-7B & 68.7 & 61.1 & - & 67.5 & 60.5 & - & - \\
        Text4Seg++ \citeinfo{(TPAMI'26)~\cite{text4segpp}}
        & Qwen2-7B & 74.8 & 64.1 & 75.8 & 73.2 & 62.8 & 74.2 & 70.8 \\
        \rowcolor{highlightgray}
        \modelP{}
        & Qwen2-2B & 73.3 & 65.3 & \underline{82.7} & 71.0 & 63.6 & \underline{81.5} & 72.9 \\
        \rowcolor{highlightgray}
        \modelP{}
        & Qwen2-7B & \underline{76.4} & \textbf{68.1} & \textbf{85.0} & \underline{75.5} & \textbf{67.0} & \textbf{84.9} & \textbf{76.2} \\
        \bottomrule
    \end{tabular}
\end{table*}

\begin{table}[!t]
    \centering
    \renewcommand{\arraystretch}{1.04}
    \footnotesize
    \caption{\textbf{Geospatial pixel reasoning results on the EarthReason benchmark.}
    Avg. is computed over validation and test gIoU/cIoU.
    Our models are highlighted with a gray background; the best and second-best results in each column are shown in bold and underlined, respectively.}
    \label{tab:earthreason_results}

    \definecolor{highlightgray}{gray}{0.93}
    \definecolor{citegray}{gray}{0.55}
    \newcommand{\citeinfo}[1]{{\color{citegray}\scriptsize #1}}

    \setlength{\tabcolsep}{3.2pt}
    \begin{adjustbox}{max width=\linewidth}
    \begin{tabular}{@{}llrrrrr@{}}
        \toprule
        \multirow{2}{*}{\textbf{Method}} &
        \multirow{2}{*}{\textbf{LLM}} &
        \multicolumn{2}{c}{\textbf{Val}} &
        \multicolumn{2}{c}{\textbf{Test}} &
        \multirow{2}{*}{\textbf{Avg.}} \\
        \cmidrule(lr){3-4} \cmidrule(lr){5-6}
        & & gIoU & cIoU & gIoU & cIoU & \\
        \midrule
        LISA (FT) \citeinfo{(CVPR'24)~\cite{lai2024lisa}}
        & Vicuna-7B & 61.0 & 57.4 & 60.9 & 59.1 & 59.6 \\
        PixelLM (FT) \citeinfo{(CVPR'24)~\cite{ren2024pixellm}}
        & Vicuna-7B & 57.9 & 57.8 & 60.0 & 59.2 & 58.7 \\
        SegEarth-R1 (FT) \citeinfo{(arXiv'25)~\cite{li2025segearth}}
        & Phi-1.5-1.3B & 68.6 & 64.1 & 70.8 & 68.3 & 68.0 \\
        Text4Seg++ \citeinfo{(TPAMI'26)~\cite{text4segpp}}
        & Qwen2-7B & 71.9 & 69.8 & 73.0 & 65.6 & 70.1 \\
        \rowcolor{highlightgray}
        \modelP{}
        & Qwen2-2B & \underline{73.3} & \underline{71.1} & \underline{74.3} & \underline{71.2} & \underline{72.5} \\
        \rowcolor{highlightgray}
        \modelP{}
        & Qwen2-7B & \textbf{74.2} & \textbf{73.3} & \textbf{75.6} & \textbf{72.8} & \textbf{74.0} \\
    \bottomrule
    \end{tabular}
    \end{adjustbox}
\end{table}

\subsection{Open-Vocabulary and Instance-Aware Segmentation}
\label{sec:exp_structured_seg}

We next evaluate the new segmentation capabilities enabled by \modelP{}.
Unlike the referring and reasoning benchmarks above, which merge all target regions into one foreground class, the following tasks require different semantic categories or object instances to remain distinguishable in a single prediction.
All results are produced by the same unified \modelP{} checkpoints evaluated in the preceding experiments.

\subsubsection{Open-Vocabulary Semantic Segmentation}

We train \modelP{} with dense semantic supervision from COCO-Stuff~\cite{cocostuff} and evaluate open-vocabulary transfer on ADE20K-150~\cite{zhou2019semantic}, Pascal Context-59~\cite{mottaghi2014role}, and Pascal VOC-20~\cite{everingham2010pascal}.
Following prior work, performance is measured by mIoU on each benchmark.
The comparison includes specialized segmentors and MLLM-based segmentation methods.

\noindent\textbf{Evaluation on ADE20K, Pascal Context, and Pascal VOC.}
These benchmarks contain different semantic vocabularies and scene distributions, testing whether instruction-defined IDs transfer beyond the COCO-Stuff training taxonomy.
As shown in \cref{tab:structured_segmentation_results}(a), \modelP{}-2B and \modelP{}-7B obtain average mIoUs of 61.3 and 63.7, respectively.
The 7B model achieves the strongest results on ADE20K-150 and Pascal Context-59 and the highest average among the compared methods.
These results show that the instruction-defined target IDs can support simultaneous semantic prediction over multiple open-vocabulary categories, rather than only one foreground/background target.

\begin{table*}[!t]
    \centering
    \renewcommand{\arraystretch}{1.05}
    \footnotesize
    \caption{\textbf{Open-vocabulary semantic and instance-aware segmentation.}
    \textbf{(a)} mIoU on ADE20K-150, Pascal Context-59, and Pascal VOC-20.
    \textbf{(b)} Validation/test gIoU and cIoU on the MUSE multi-target reasoning benchmark.
    Avg. is computed over the three mIoUs in (a) and the four validation/test scores in (b).
    Our models are highlighted with a gray background; bold and underlined denote the best and second-best results in each column, respectively.}
    \label{tab:structured_segmentation_results}

    \definecolor{highlightgray}{gray}{0.93}
    \definecolor{citegray}{gray}{0.55}
    \newcommand{\citeinfo}[1]{{\color{citegray}\footnotesize #1}}

    \setlength{\tabcolsep}{9pt}
    \begin{tabular}{@{}llcccc@{}}
        \toprule
        \multicolumn{6}{@{}l}{\textbf{(a) Open-vocabulary semantic segmentation}} \\
        \midrule
        \textbf{Method} & \textbf{LLM} & \textbf{ADE-150} & \textbf{PC-59} & \textbf{PAS-20} & \textbf{Avg.} \\
        & & \textbf{mIoU} & \textbf{mIoU} & \textbf{mIoU} & \\
        \midrule
        ClearCLIP \citeinfo{(ECCV'24)~\cite{clearclip}} & -- & 16.7 & 35.9 & 80.9 & 44.5 \\
        ProxyCLIP \citeinfo{(ECCV'24)~\cite{proxyclip}} & -- & 24.2 & 39.6 & 83.3 & 49.0 \\
        MaskCLIP \citeinfo{(ICML'23)~\cite{maskclip}} & -- & 23.7 & 45.9 & -- & -- \\
        GroupViT \citeinfo{(CVPR'22)~\cite{groupvit}} & -- & 9.2 & 23.4 & 79.7 & 37.4 \\
        OVSeg \citeinfo{(CVPR'23)~\cite{ovseg}} & -- & 24.8 & 53.3 & \underline{92.6} & 56.9 \\
        SAN \citeinfo{(TPAMI'23)~\cite{san}} & -- & 27.5 & 53.8 & \textbf{94.0} & 58.4 \\
        LaSagnA \citeinfo{(arXiv'24)~\cite{lasagna}} & Vicuna-7B & 14.3 & 46.1 & 69.8 & 43.4 \\
        Text4Seg \citeinfo{(ICLR'25)~\cite{lan2024text4seg}} & Vicuna-7B & 16.5 & 52.5 & 76.5 & 48.5 \\
        \rowcolor{highlightgray}
        \modelP{} & Qwen2-2B & \underline{41.0} & \underline{55.4} & 87.6 & \underline{61.3} \\
        \rowcolor{highlightgray}
        \modelP{} & Qwen2-7B & \textbf{42.2} & \textbf{56.9} & 92.1 & \textbf{63.7} \\
        \bottomrule
    \end{tabular}

    \vspace{0.7em}

    \setlength{\tabcolsep}{8pt}
    \begin{tabular}{@{}llrrrrr@{}}
        \toprule
        \multicolumn{7}{@{}l}{\textbf{(b) Instance-aware multi-target reasoning segmentation on MUSE}} \\
        \midrule
        \multirow{2}{*}{\textbf{Method}} &
        \multirow{2}{*}{\textbf{LLM}} &
        \multicolumn{2}{c}{\textbf{Val}} &
        \multicolumn{2}{c}{\textbf{Test}} &
        \multirow{2}{*}{\textbf{Avg.}} \\
        \cmidrule(lr){3-4} \cmidrule(lr){5-6}
        & & gIoU & cIoU & gIoU & cIoU & \\
        \midrule
        LISA \citeinfo{(CVPR'24)~\cite{lai2024lisa}}
        & Vicuna-7B & 17.2 & 28.8 & 24.4 & 36.5 & 26.7 \\
        GSVA \citeinfo{(CVPR'24)~\cite{xia2024gsva}}
        & Vicuna-7B & 38.9 & 40.9 & 44.3 & 54.1 & 44.6 \\
        PixelLM \citeinfo{(CVPR'24)~\cite{ren2024pixellm}}
        & Vicuna-7B & 41.9 & 48.9 & 44.0 & 57.8 & 48.2 \\
        POPEN \citeinfo{(CVPR'25)~\cite{popen}}
        & Vicuna-7B & 45.4 & 55.2 & 46.4 & 62.9 & 52.5 \\
        Text4Seg++ \citeinfo{(TPAMI'26)~\cite{text4segpp}}
        & Qwen2-7B & \textbf{70.4} & 57.7 & 63.2 & 63.8 & \textbf{63.8} \\
        \rowcolor{highlightgray}
        \modelP{}
        & Qwen2-2B & 59.1 & \underline{58.6} & \underline{63.4} & \underline{64.1} & 61.3 \\
        \rowcolor{highlightgray}
        \modelP{}
        & Qwen2-7B & \underline{61.0} & \textbf{60.4} & \textbf{65.8} & \textbf{66.9} & \underline{63.5} \\
        \bottomrule
    \end{tabular}
\end{table*}

\subsubsection{Instance-Aware Segmentation}

Instance-aware segmentation goes beyond semantic segmentation by requiring individual objects, including those from the same category, to remain distinguishable in the output.
We evaluate this capability on MUSE~\cite{ren2024pixellm}, where the model must infer the intended object instances from reasoning-intensive instructions and segment each instance separately.
This setting directly tests whether the structured target IDs introduced by \modelP{} can support instance-level reasoning and prediction.
We report gIoU and cIoU on both validation and test splits, with the overall average computed over these four scores.

\noindent\textbf{Evaluation on MUSE.}
As shown in \cref{tab:structured_segmentation_results}(b), \modelP{}-2B and \modelP{}-7B achieve average scores of 61.3 and 63.5.
The 7B model is competitive with Text4Seg++ overall and obtains higher gIoU and cIoU on the test split (65.8/66.9 vs. 63.2/63.8).
These results demonstrate a substantial extension from \model{}'s single-target binary-mask prediction to \modelP{}'s structured multi-instance prediction.
Although \model{} could in principle recover multiple instance masks through repeated binary predictions, doing so requires a separate, unambiguous referring expression for each instance and becomes impractical in complex scenes.
In contrast, \modelP{} can discover and distinguish multiple instances from a single instruction and return them jointly in one structured prediction.

\noindent\textbf{Qualitative Results.}
\cref{fig:stamp_showcase,fig:stamplus_showcase} illustrate the progression from single-target binary masks to small-target, open-vocabulary, instance-aware, and panoptic-style predictions.

\subsection{Segmentation--Understanding Interaction}
\label{sec:exp_seg_understanding}

Beyond the primary segmentation evaluations, we conduct two analysis experiments to better understand the benefits and broader potential of the structured extension in \modelP{}.
We examine whether more precise Phase~1 target descriptions improve Phase~2 segmentation and whether the resulting spatial grounding capabilities can support downstream visual understanding.
These analyses thus help explain the value of extending \model{} beyond single-target binary-mask prediction to a structured multi-target interface.

\subsubsection{Phase~1-Assisted Segmentation}

During Phase~1, \modelP{} describes each target by generating its textual label and target ID, along with an optional \texttt{bbox} indicating its location.
In the preceding segmentation experiments, the bounding boxes in this field are generated by the model itself during Phase~1.
The two-phase interface also allows us to augment the Phase 1 target specification with a human-provided bounding box before Phase 2, analogous to providing a box prompt to SAM.
This controlled setting allows us to analyze how the accuracy of the spatial target description affects subsequent mask prediction.

\noindent\textbf{Evaluation with Human-Provided Target Cues.}
As shown in \cref{tab:seg_understanding_results}(a), \modelP{}-2B improves from an average cIoU of 77.4 to 89.1 when supplied with the target boxes.
This large gain shows that the quality of the Phase~1 target description has a substantial impact on Phase~2 segmentation.
It also suggests a clear path for further improvement: Phase~1 could be strengthened through reinforcement-learning-based optimization or assistance from specialized localization tools, with the resulting spatial cues directly benefiting Phase~2.
This form of modular enhancement is enabled by the structured design of \modelP{}, which exposes an explicit and editable intermediate target description before mask prediction.

\begin{table}[!t]
    \centering
    \renewcommand{\arraystretch}{1.04}
    \footnotesize
    \caption{\textbf{Segmentation--understanding interaction analyses.}
    \textbf{(a)} Augmenting the Phase 1 target specification with a human-provided box improves referring segmentation.
    \textbf{(b)} Segmentation-derived spatial grounding supports look-twice reasoning on multimodal benchmarks.
    Our models are highlighted with a gray background; bold and underlined denote the best and second-best results, respectively.}
    \label{tab:seg_understanding_results}

    \setlength{\tabcolsep}{2.4pt}
    \begin{tabular}{@{}lrrrr@{}}
        \toprule
        \multicolumn{5}{@{}l}{\textbf{(a) Human-provided target cues for referring segmentation}} \\
        \midrule
        \textbf{Method} & \textbf{RefCOCO} & \textbf{RefCOCO+} & \textbf{RefCOCOg} & \textbf{Avg.} \\
        & testB & testB & test(U) & \\
        \midrule
        LISA~\cite{lai2024lisa} & 72.3 & 58.1 & 70.6 & 67.0 \\
        GSVA~\cite{xia2024gsva} & 73.5 & 59.8 & 73.3 & 68.9 \\
        Text4Seg~\cite{lan2024text4seg} & 76.2 & 66.1 & 73.9 & 72.1 \\
        Text4Seg++~\cite{text4segpp} & 78.9 & 70.9 & 78.9 & 76.2 \\
        \rowcolor{highlightgray}
        \model{}-2B & \underline{79.5} & 72.7 & 78.8 & 77.0 \\
        \rowcolor{highlightgray}
        \modelP{}-2B & 79.4 & \underline{73.1} & \underline{79.6} & \underline{77.4} \\
        \rowcolor{highlightgray}
        \modelP{}-2B w/ bbox & \textbf{89.1} & \textbf{89.3} & \textbf{88.9} & \textbf{89.1} \\
        \bottomrule
    \end{tabular}

    \vspace{0.7em}

    \setlength{\tabcolsep}{2.1pt}
    \begin{tabular}{@{}lrrrrr@{}}
        \toprule
        \multicolumn{6}{@{}l}{\textbf{(b) Look-twice reasoning on multimodal benchmarks}} \\
        \midrule
        \textbf{Model / Mode} & \textbf{TextVQA} & \textbf{InfoVQA} & \textbf{POPE} & \textbf{DocVQA} & \textbf{Avg.} \\
        \midrule
        Qwen2-VL-2B / Direct & 70.7 & 38.9 & 85.2 & 70.9 & 66.4 \\
        \rowcolor{highlightgray}
        \modelP{}-2B / Direct & {70.6} & 40.2 & 85.4 & 71.3 & 66.9  \\
        Qwen2-VL-2B / Look-twice & \underline{74.0} & \underline{41.3} & \underline{87.2} & \underline{71.6} & \underline{68.5} \\
        \rowcolor{highlightgray}
        \modelP{}-2B / Look-twice & \textbf{74.7} & \textbf{46.3} & \textbf{87.5} & \textbf{75.7} & \textbf{71.1} \\
        \bottomrule
    \end{tabular}
\end{table}

\subsubsection{Segmentation-Assisted Understanding}

Segmentation training provides \modelP{} with fine-grained spatial grounding, and we test whether it benefits downstream visual understanding. Because the original Qwen2-VL training data are unavailable, both models use the same released Qwen2-VL-2B initialization and LLaVA-665k visual-instruction data; only \modelP{} receives additional segmentation data.

We follow the look-twice protocol~\cite{zhang2025mllms}, deriving question-relevant regions from first-pass attention to guide a second answer-generation pass. We use attention rather than predicted masks because it preserves continuous spatial evidence without introducing mask-classifier errors, and \modelP{} is not trained to feed discrete masks back into visual question answering. The experiment therefore tests whether segmentation-learned grounding is reflected in reusable internal attention.

\noindent\textbf{Evaluation with Look-Twice Reasoning.}
As reported in \cref{tab:seg_understanding_results}(b), applying the look-twice protocol to Qwen2-VL-2B raises its average score from 66.4 to 68.5, showing that a second, spatially guided pass is beneficial even without segmentation training.
Under the same inference protocol, \modelP{}-2B achieves an average score of 71.1 across TextVQA, InfoVQA, POPE, and DocVQA, exceeding Qwen2-VL-2B by 2.6 points and outperforming it on every benchmark.
The results therefore provide preliminary evidence that structured segmentation training produces more informative internal spatial attention that can be reused for downstream visual understanding.

\subsection{General Properties and Analysis}
\label{sec:exp_general_efficiency}
\label{sec:exp_ablation}

The preceding experiments have evaluated the segmentation accuracy and task coverage of \model{} and \modelP{}, ranging from single-target referring and reasoning segmentation to remote-sensing small-target, open-vocabulary, and instance-aware prediction, as well as their interaction with downstream visual understanding.
To complete the evaluation of the accuracy--dialogue-compatibility--efficiency trilemma, we next test whether both models preserve general multimodal ability, analyze inference efficiency in the single-target and structured multi-target regimes, and ablate the shared All-Mask components.

\subsubsection{General Multimodal Ability}

Following Text4Seg~\citep{lan2024text4seg}, we compare segmentation-only training with mixed training that combines the corresponding segmentation data with the LLaVA-665k visual-instruction set.
We evaluate segmentation on the RefCOCO-family validation sets and general multimodal ability on MMMU~\cite{mmmu}, MMBench~\cite{liu2024mmbench}, MMStar~\cite{mmstar}, ScienceQA~\citep{lu2022learn}, TextVQA~\citep{singh2019towards}, and VizWiz~\citep{gurari2018vizwiz}.

\noindent\textbf{Evaluation on General Multimodal Benchmarks.}
As shown in \cref{tab:vqa_res_seamless}, mixed training allows \model{}-2B to retain performance close to the Qwen2-VL-2B reference across the general benchmarks while simultaneously acquiring strong segmentation ability.
It substantially avoids the dialogue collapse observed for some embedding-prediction baselines and improves RES performance over segmentation-only training.
\modelP{}-2B similarly maintains strong multimodal results while improving average RES performance, showing that the structured extension does not trade away the dialogue compatibility established by \model{}.
This verifies the dialogue-compatibility side of the broader trilemma: expanding from binary target masks to structured multi-target prediction does not compromise general multimodal instruction following.

\begin{table*}[!t]
    \centering
    \renewcommand{\arraystretch}{1.05}
    \footnotesize

    \caption{
       \textbf{Joint visual understanding and segmentation performance.}
       LLaVA-1.5-7B and Qwen2-VL-2B fine-tuned on LLaVA-665k serve as VQA references.
       Training uses segmentation data (Seg.), visual-instruction data (VQA), or their mixture (Mix).
       RES columns report cIoU on the RefCOCO-family validation sets.
       Reference models and the segmentation-only row without VQA scores are shown in gray text; our models are highlighted with a gray background.
       Bold and underlined denote the best and second-best results in each column, respectively.
    }
    \label{tab:vqa_res_seamless}
    
    \definecolor{highlightgray}{gray}{0.93}
    \definecolor{textgray}{gray}{0.45}

    \setlength{\tabcolsep}{4pt}
    \begin{tabular}{@{}ll rrrr rr ccc@{}}
        \toprule
        \multirow{2}{*}{\textbf{Methods}} & \multirow{2}{*}{\textbf{Training Data}} & \multicolumn{6}{c}{\textbf{VQA}} & \multicolumn{3}{c}{\textbf{RES (val)}} \\
        \cmidrule(lr){3-8} \cmidrule(lr){9-11}
        & & MMMU & MMBench & MMStar & ScienceQA & TextVQA &  VizWiz & RefC & RefC+ & RefCg \\
        \midrule
        \textcolor{textgray}{LLaVA-1.5-7B} & \textcolor{textgray}{VQA} & \textcolor{textgray}{35.7} & \textcolor{textgray}{66.5} & \textcolor{textgray}{33.1} & \textcolor{textgray}{68.4} & \textcolor{textgray}{55.0} & \textcolor{textgray}{50.0} & \textcolor{textgray}{n.a.} & \textcolor{textgray}{n.a.} & \textcolor{textgray}{n.a.} \\

        \textcolor{textgray}{Qwen2-VL-2B} & \textcolor{textgray}{VQA} & \textcolor{textgray}{\underline{38.3}} & \textcolor{textgray}{66.5} & \textcolor{textgray}{42.1} & \textcolor{textgray}{70.2} & \textcolor{textgray}{\textbf{70.7}} & \textcolor{textgray}{\textbf{60.3}} & \textcolor{textgray}{n.a.} & \textcolor{textgray}{n.a.} & \textcolor{textgray}{n.a.} \\
        LISA{-7B} & Mix & 0 & 0 & 0 & 0 & 0 & 0 & 74.9 & 65.1 & 67.9 \\
        READ{-7B} & Mix & 1.1 & 0 & 14.4 & 23.2 & 22.6 & 1.3 & 78.1 & 68.4 & 70.1 \\
        Text4Seg{-7B} & Mix & 34.0 & 54.8 & 33.4 & 68.1 & 55.0 & 50.9 & 79.3 & 72.1 & 72.1 \\
        [3pt]
        \rowcolor{highlightgray}
         \textcolor{textgray}{\model{-2B}} & \textcolor{textgray}{Seg.} & \textcolor{textgray}{n.a.} & \textcolor{textgray}{n.a.} & \textcolor{textgray}{n.a.} & \textcolor{textgray}{n.a.} & \textcolor{textgray}{n.a.} & \textcolor{textgray}{n.a.} & \textcolor{textgray}{81.9} & \textcolor{textgray}{77.1} & \textcolor{textgray}{78.5} \\
        \rowcolor{highlightgray}
        \model{-2B} & Mix & 37.8 & \underline{68.7} & \underline{42.4} & \underline{72.6} & 69.7 & \underline{59.9} & \textbf{82.2} & \underline{77.3} & \underline{79.0} \\
        \rowcolor{highlightgray}
        \modelP{-2B} & Mix & \textbf{39.3} & \textbf{68.9} & \textbf{42.9} & \textbf{74.4} & \underline{70.6} & 59.2 & \underline{82.0} & \textbf{78.3} & \textbf{79.4} \\
        \bottomrule
    \end{tabular}
\end{table*}

\subsubsection{Efficiency Analysis}

\begin{figure}[t]
    \centering
    \includegraphics[width=0.9\linewidth]{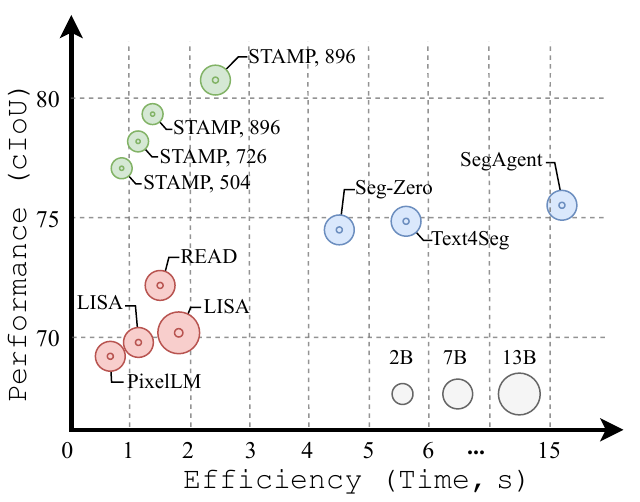}
    \caption{\textbf{Efficiency comparison across MLLM segmentation paradigms.}
    Methods from the same paradigm are grouped by color; numbers following model names denote input resolution, and marker size indicates model scale.
    Higher and farther left is better. Latency is measured on a single NVIDIA A800 using the same test case for all methods.}
    \label{fig:speed}
\end{figure}

\noindent\textbf{Accuracy--Latency Trade-off.}
We first revisit the single-target segmentation regime that motivated \model{}.
As illustrated in \cref{fig:speed}, \model{} achieves a favorable accuracy--latency trade-off.
Its mask-generation latency is comparable to efficient embedding-prediction approaches and substantially lower than autoregressive next-token methods that generate long mask sequences.
The resolution study in \cref{tab:allmask_ablation} further shows that \model{} can trade a small amount of accuracy for lower latency without retraining.

\begin{figure}[t]
    \centering
    \includegraphics[width=0.95\linewidth]{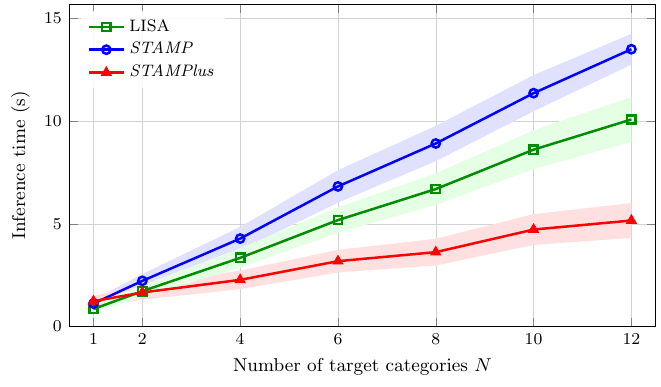}
    \caption{\textbf{Inference-time scaling with the number of target categories.}
    \model{} and LISA repeat a separate prediction of similar cost for each category, resulting in a steep latency increase.
    The additional cost of \modelP{} is concentrated in the Phase~1 autoregressive generation of the structured target list, while all categories share a single Phase~2 mask prediction, yielding a substantially smaller slope.
    Lines show mean latency and shaded bands show the corresponding standard deviation; all measurements use a single NVIDIA H800 GPU.
    }
    \label{fig:target_scaling_speed}
\end{figure}

\noindent\textbf{Multi-Target Inference.}
The preceding analysis establishes the efficiency advantage of \model{} in the single-target setting; we next examine the multi-target setting, where several semantic categories must remain separate.
Single-target binary-mask methods such as \model{} and LISA require one target-specific prediction per category, whereas \modelP{} assigns all target IDs in one structured mask map.
As shown in \cref{fig:target_scaling_speed}, we compare the inference-time scaling of LISA, \model{}, and \modelP{}.
The evaluation uses identical image buckets and target-category counts under the same hardware and input-resolution setting.

These results verify the efficiency side of the broader trilemma: although \modelP{} supports multiple target identities, its Phase~2 mask generation remains a single shared forward pass and scales substantially better than repeated target-specific prediction.
Combined with the preceding accuracy and dialogue-compatibility results, this completes the evidence that \modelP{} resolves the segmentation trilemma in broader segmentation settings.

\subsubsection{Ablation and Generalization Analysis}

We finally distinguish the shared All-Mask foundation from the adaptations specific to \modelP{}.

\noindent\textbf{Shared All-Mask Components.}
As shown in \cref{tab:allmask_ablation}, removing either the visually enhanced mask embedding $\mathbf{E}_{\mathrm{mask}}$ or the hybrid attention mask $\mathbf{A}_{\mathrm{hyb}}$ leads to a clear accuracy drop.
This verifies that patch-aligned visual initialization and bidirectional interaction among mask tokens both contribute to dense prediction quality.

\noindent\textbf{Resolution Flexibility.}
Although \model{} is trained primarily at a resolution of $896\times896$, it can directly process lower-resolution inputs.
Reducing the resolution to $726\times726$ or $504\times504$ lowers inference time while retaining competitive cIoU, providing a controllable accuracy--efficiency trade-off.

\begin{table}[!t]
\centering
\footnotesize
\renewcommand{\arraystretch}{1.06}
\caption{\textbf{Ablation and input-resolution study of \model{}.} We report average RefCOCO-family cIoU and inference time measured on a single NVIDIA A800.}
\label{tab:allmask_ablation}
\setlength{\tabcolsep}{5.5pt}
\begin{tabular}{@{}llrr@{}}
\toprule
\textbf{Method} & \textbf{Input Size} & \textbf{cIoU} & \textbf{Time (s)} \\
\midrule
\model{-2B} & $896 \times 896$ & \underline{79.1} & 1.3 \\
\quad w/o $\mathbf{A}_{\text{hyb}}$ & $896\times896$ & 76.2 & 1.3 \\
\quad w/o $\mathbf{E}_{\text{mask}}$ & $896\times896$ & 73.3 & 1.3 \\
\quad w/o SAM & $896\times896$ & 75.3 & 0.9 \\
\model{-2B} & $726 \times 726$ & 78.6 & 1.1 \\
\model{-2B} & $504 \times 504$ & 77.1 & 0.9 \\
\model{-7B} & $896 \times 896$ & \textbf{80.7} & 2.4 \\
\bottomrule
\end{tabular}
\end{table}

\noindent\textbf{High-Resolution Scaling for Small Targets.}
At the standard natural-image resolution, a small remote-sensing target may be represented by only a few visual patches.
Increasing the input resolution retains more patch features over the target, but it also enlarges the patch grid; since every retained patch is paired with one image-aligned mask token, the mask-token budget must grow accordingly to preserve this finer spatial sampling.
We therefore use higher-resolution inputs and increase the budget from 1024--1280 to 2560--3200 for the remote-sensing experiments in \cref{tab:rrsis_d_results,tab:earthreason_results}.
To isolate the effect of this scaling choice, \cref{tab:dynamic_resolution_ablation} uses the same checkpoint and evaluation images while jointly scaling the input resolution and mask-token budget on RRSIS-D and EarthReason.

\begin{table}[!t]
    \centering
    \renewcommand{\arraystretch}{1.04}
    \footnotesize
    \caption{\textbf{Controlled input-resolution and mask-token scaling.}
    Using the same \modelP{}-2B checkpoint and evaluation images, the 1024--1280 and 2560--3200 token budgets correspond to standard- and high-resolution inputs, respectively; results are reported on the test splits.}
    \label{tab:dynamic_resolution_ablation}

    \setlength{\tabcolsep}{2.5pt}
    \begin{tabular}{@{}lrrrrr@{}}
        \toprule
        \multirow{2}{*}{\textbf{Mask-token Budget}} &
        \multicolumn{3}{c}{\textbf{RRSIS-D}} &
        \multicolumn{2}{c}{\textbf{EarthReason}} \\
        \cmidrule(lr){2-4} \cmidrule(lr){5-6}
        & Acc@0.5 & gIoU & cIoU & gIoU & cIoU \\
        \midrule
        1024--1280 &  63.2 & 58.1 & 76.5 & 69.4 & 70.7\\
        2560--3200 & 71.0 & 63.6 & 81.5 & 74.3 & 71.2 \\
        \bottomrule
    \end{tabular}
\end{table}

Jointly increasing the input resolution and mask-token budget improves gIoU by 5.5 points on RRSIS-D and 4.9 points on EarthReason.
This metric is particularly sensitive to small-target quality because it averages IoU over samples, giving a small object the same sample-level weight as a large one; even a minor boundary or localization error can cover a substantial fraction of a small target.
The denser patch--token grid therefore produces a pronounced gIoU gain by preserving fine small-object evidence, whereas cIoU aggregates intersections and unions over the dataset and can be dominated by larger regions.

\noindent\textbf{Backbone Transfer.}
Replacing Qwen2-VL with LLaVA (Vicuna-7B) yields an average RefCOCO-family cIoU of 77.0 in \cref{tab:final_with_citations_sorted}.
Although below the Qwen2-VL variants, the LLaVA-based model still surpasses every compared prior method with a reported average except Text4Seg++ (78.9).

\FloatBarrier

\section{Conclusion}
We introduced All-Mask Prediction, which decouples autoregressive dialogue generation from non-autoregressive dense mask prediction within the native MLLM token interface.
Its binary instantiation, \emph{Binary All-Mask Prediction}, is implemented by \model{} and resolves the segmentation trilemma for task-specific single-target segmentation.
We further generalized this paradigm to \emph{Structured All-Mask Prediction} and developed \modelP{}, which dynamically binds the target identities generated in Phase~1 to a shared structured mask space in Phase~2.
Using one unified checkpoint without task-specific fine-tuning, \modelP{} preserves the referring and reasoning segmentation ability of \model{} while extending to remote-sensing small-target, open-vocabulary semantic, and instance-aware segmentation; a qualitative panoptic-style example further illustrates the extensibility of the structured interface.
Across these broader settings, our experiments establish all three sides of the trilemma: strong segmentation performance, preserved general multimodal instruction-following ability, and efficient mask generation through a single shared Phase~2 pass.
Thus, \modelP{} extends the resolution of the segmentation trilemma from task-specific single-target segmentation to broader segmentation settings.
Beyond task expansion, our analyses reveal how structured Phase~1 target generation guides Phase~2 mask prediction and indicate that segmentation-learned spatial grounding can support downstream multimodal understanding.
Overall, All-Mask Prediction provides a general and scalable interface for accurate, dialogue-compatible, and efficient dense perception.

\bibliographystyle{IEEEtran}
\bibliography{main}

\end{document}